\documentclass[11pt]{article} % For LaTeX2e
\usepackage[final]{acl}
\usepackage{amsmath, amssymb, amsthm, mathtools}
\usepackage{multirow}
\usepackage{algorithm}
\usepackage{dsfont}
\usepackage{algpseudocode}
\usepackage{listings}
\usepackage{hyperref}
\usepackage{tabularray}
\usepackage{xcolor}
\usepackage{enumitem}
\usepackage{microtype}
\usepackage{hyperref}
\usepackage{url}
\usepackage{supertabular}
\usepackage{cleveref}
\usepackage{booktabs}
\usepackage{subcaption}
\usepackage{natbib}
\usepackage{booktabs,tabularx,array,longtable,makecell}
\usepackage{times}
\usepackage{inconsolata}
\usepackage{latexsym}
\usepackage[T1]{fontenc}
\usepackage[utf8]{inputenc}
\usepackage{microtype}
\usepackage{graphicx}

\title{Test-Time Scaling for Scientific Equation Discovery}

\author{
  \textbf{Haowei Lin}\thanks{~~Equal Contribution. Code and data are available at \url{https://ahong-lin.github.io/ScaleSR-tts/}.} \quad
  \textbf{Hubert Lim}\footnotemark[1] \\
  \textbf{Xiangyu Wang} \quad
  \textbf{Letian Huang} \quad
  \textbf{Di He} \\
  \addlinespace
  Peking University \\
  \texttt{\{linhaowei, di\_he\}@pku.edu.cn}, \quad \texttt{HubertLinHong@stu.pku.edu.cn}
}

\begin{document}

\maketitle

\begin{abstract}
Test-time scaling (TTS) improves language model reasoning by allocating additional test-time compute, but prior work mainly studies closed-ended tasks such as math and coding. We study TTS for automated equation discovery, an open-ended setting where models search over candidate equations and rely on observed datapoints for feedback. We formulate LLM-driven equation discovery as an iterative search process that unifies Best-of-$N$, sequential refinement, tree search, and evolution-style methods under a common compute-allocation view. To isolate allocation effects from prompt engineering and other heuristics, we compare minimal parallel controllers under fixed budgets. On LLM-SRBench equation-discovery tasks, we find that search width is the dominant allocation parameter: the best width in our sweep generally increases with the compute budget, while the population--branching split and controller choice matter less. Appropriate width selection also improves wall-clock efficiency by increasing parallelism. These results suggest that, given an informative verifier, controlling exploration and exploitation is central to scaling LLM-based equation discovery.
\end{abstract}

\section{Introduction}

Test-time scaling (TTS) has become a standard way to improve the reasoning performance of language models by allocating additional compute at inference time rather than updating model parameters \citep{song2025good,snell2024scaling}. Most prior work studies closed-ended reasoning tasks such as mathematics and code generation, where the objective is to recover a correct answer to a well-specified problem \citep{guan2025rstar,li2025sstar}. In these settings, TTS is often framed as a question of \emph{how much} extra inference-time compute to spend~\citep{snell2024scaling}, or how more compute can help smaller models approach the performance of larger ones~\citep{wu2024inference}.

In this paper, we study TTS for automated equation discovery. Equation discovery is narrower than scientific discovery as a whole, but it captures an important open-ended modeling problem: a system must propose candidate equations or programs, receive feedback from an external evaluator, and iteratively search for better candidates. Unlike closed-ended reasoning benchmarks, there is typically no single target string to recover during search; progress is measured by improvement under a task-specific verifier. This makes equation discovery a useful testbed for studying how inference-time compute should be allocated across exploration and refinement.

This shift in perspective matters because recent LLM-based discovery and program-search systems already hint that the answer is not obvious. A growing line of work uses language models to generate candidate artifacts and closes the loop with an external verifier \citep{romeraparedes2024mathematical,novikov2025alphaevolve,lange2025shinkaevolve}. These systems often work well, but they are usually introduced as complicated agent workflows with many coupled design choices, including specialized prompts, mutation rules, model ensembles, and hand-tuned pruning heuristics. As a result, it is difficult to tell which ingredients are fundamental. In particular, the underlying \emph{control flow} of test-time search is often entangled with domain-specific engineering.

We argue that, for LLM-based equation discovery, compute allocation is a first-order design choice. Once an informative verifier is available, different search procedures can be understood primarily as different ways of allocating a fixed test-time budget across exploration and exploitation. This viewpoint places classical TTS methods and recent evolution-style systems in a common framework. Best-of-$N$~\citep{song2025good} corresponds to one-shot wide exploration, sequential refinement~\citep{madaan2023self,chen2023teaching} corresponds to narrow multi-step exploitation, tree search~\citep{chen2024tree} corresponds to structured branching with repeated pruning, and evolution-style systems~\citep{novikov2025alphaevolve,lange2025shinkaevolve} correspond to persistent multi-candidate or multi-island search with repeated selection and expansion. Under this view, the central object of study is not a particular agent scaffold, but the control law governing which candidates to expand, how many continuations to generate, and how aggressively to retain or discard candidates over time.

To study this question, we formulate LLM-driven equation discovery as an iterative search process with four generic operations: \emph{sample}, \emph{generate}, \emph{evaluate}, and \emph{prune}. This abstraction isolates compute allocation from domain-specific heuristics and lets us compare different control flows under a shared budget. We then instantiate the framework with two simple, highly parallel controllers: \emph{Parallel Beam Search} (PBeam), which expands the current frontier and keeps the strongest candidates, and \emph{Parallel Iterative Expansion} (PIE), which instead guides future expansions from the full search history. These controllers are intentionally minimal. Their purpose is not to encode human-crafted priors, but to provide a clean testbed for asking which allocation decisions matter.

Our experiments on LLM-SRBench automatic equation discovery reveal a simple and consistent picture. The dominant control variable is \emph{search width}. Neither greedy exploitation nor one-shot broad exploration performs best. Instead, the best-performing width lies in the interior and shifts upward as the total test-time budget increases. This same allocation choice also improves wall-clock efficiency: moderate width exposes more parallelism, so better search quality need not come at the cost of slower runtime. By contrast, after width is chosen, the exact split between population size and branching factor, as well as the choice between PBeam and PIE, has a much smaller effect.

Our goal is therefore not to propose a universal controller for scientific discovery, nor to claim that equation discovery fully represents all scientific-discovery problems. Rather, we use equation discovery as a controlled benchmark setting to show that \emph{control flow itself} is an important object of study for open-ended LLM search. Given an informative evaluator, there is substantial empirical structure in how a fixed compute budget should be spent. Making that structure explicit yields both a unified language for comparing existing equation-discovery systems and a practical recipe for scaling this class of test-time search problems.

\paragraph{Contributions.}
\begin{itemize}[leftmargin=*,nosep]
    \item We formulate LLM-driven equation discovery as a unified external TTS process that subsumes common control flows including Best-of-$N$, sequential refinement, tree search, and recent evolution-style systems.
    \item On LLM-SRBench automated equation discovery, we show that search width is the dominant allocation parameter. In our sweep, larger budgets tend to favor wider searches, and optimizing this trade-off can improve both final performance and wall-clock efficiency.
    \item Our analysis is based on a large-scale controlled empirical study spanning roughly 4,000 H100 GPU hours, and we will release our code, results, prompts, and evaluation framework to support future research on TTS for equation discovery.
\end{itemize}

\subsection{Background: TTS for equation discovery}

\paragraph{Overview of TTS} TTS refers to improving an LLM's performance at inference time by allocating additional computation without updating model parameters \citep{welleck2024meta,snell2024scaling}. TTS has shown strong gains on reasoning-intensive domains, especially mathematics and code generation \citep{wang2023selfconsistency,brown2024monkeys,li2025sstar,snell2024scaling}. In this paper, we focus on \emph{external} TTS, where extra compute is realized through multiple model calls. We use \emph{internal} TTS to denote the complementary regime where a single response is allowed to consume more reasoning tokens.

\paragraph{External TTS} For clarity, we organize external TTS by control flow. \emph{Parallel} scaling generates multiple candidates independently and aggregates them afterward, e.g., via self-consistency, repeated sampling, or verifier-based Best-of-$N$ selection \citep{wang2023selfconsistency,brown2024monkeys,cobbe2021trainingverifiers}.  \emph{Sequential} scaling iteratively refines intermediate outputs, conditioning subsequent generation steps on prior states to enable self-correction \citep{gou2023critic}.\emph{Hybrid} scaling combines branching with sequential refinement, yielding explicit search over candidate states rather than a single left-to-right rollout, thereby enabling iterative revision, look-ahead, or backtracking \citep{yao2023tree,li2025sstar}.

\paragraph{Verification} The performance of TTS depends critically on \emph{verification}. In benchmark reasoning tasks, verification is often implemented by learned verifiers or reward models: outcome reward models score final answers, whereas process reward models evaluate intermediate reasoning steps \citep{cobbe2021trainingverifiers,lightman2023verify}. These signals can be used either to guide search online or to rerank candidate solutions offline.

\paragraph{TTS in scientific discovery} Recent work has used an LLM to propose candidate programs or constructions, while an external evaluator (e.g., mathematical objective) assigns fitness and closes the search loop \citep{romeraparedes2024mathematical,novikov2025alphaevolve}. This discovery setting differs from standard math or coding benchmarks in an important respect: the objective is typically \emph{open-ended}. There is often no unique ground-truth answer known in advance; instead, progress is measured by improvement under a task-specific evaluator. When that evaluator is reliable, the main bottleneck shifts from recovering a known answer to allocating test-time compute effectively over a large search space of candidate solutions. Under this lens, recent self-evolving systems~\citep{lange2025shinkaevolve,assumpccao2025codeevolve} can be viewed as specialized instances of external TTS that differ primarily in their control flow.

\subsection{A unified TTS formulation}

We formalize LLM-driven equation discovery as an iterative search over a dynamic memory state $\mathcal{M}_t$, which maintains the candidate solutions, such as equations or executable programs, available at step $t$. $\mathcal{M}_0$ can be initialized as a naive solution to the problem. By abstracting away domain-specific details such as prompt design or mutation rules, we obtain a common computational framework governed by a global test-time compute budget $N$. Each iteration is specified by four ingredients: a selection policy over $\mathcal{M}_t$, a branching rule, a verifier, and a pruning operator.

\paragraph{Sample} Let $q_t(\cdot \mid \mathcal{M}_t)$ denote a selection policy over the current memory state. Select a subset
\begin{gather*}
S_t = \{x_t^{(i)}\}_{i=1}^{n_t} \sim q_t(\cdot \mid \mathcal{M}_t), \\
S_t \subseteq \mathcal{M}_t, \qquad |S_t| = n_t.
\end{gather*}
for expansion. The policy $q_t$ may be greedy or score-based, and determines which previously discovered candidates are revisited at step $t$.

\paragraph{Generate} For each candidate $x_t^{(i)} \in S_t$,\footnote{In practice, we can select multiple candidates to construct the context for the LLM. This results in a directed acyclic graph rather than a tree in the control flow, although the parameterization remains similar. For simplicity, we follow recent practices~\citep{wang2025thetaevolve,yuksekgonul2026learning} to focus on the single candidate setting.} query the LLM to generate a set $G_t^{(i)}$ of $k_t^{(i)}$ new proposals (e.g., refinements or novel variations). The full generated set is
\[
G_t = \bigcup_{i=1}^{n_t} G_t^{(i)},
\qquad
|G_t| = \sum_{i=1}^{n_t} k_t^{(i)}.
\]
When $k_t^{(i)} = k_t$ for all $i$, we write $k_t$ for the common branching factor.

\paragraph{Evaluate and prune} Score $G_t$ using a task-specific verifier, merge the new proposals with the current memory, and apply a pruning operator $\Pi_t$ (e.g., FIFO, top-$B$ selection or diversity filtering) to retain the surviving candidates:
\[
\mathcal{M}_{t+1} = \Pi_t(\mathcal{M}_t \cup G_t).
\]

\paragraph{Iterate} Repeat this process for $T$ steps until the global compute budget $N$ is exhausted. Assuming a uniform cost per generated proposal,\footnote{Traditional TTS literature use FLOPs as their compute measure, which is mainly affected by the number of model rollouts (i.e., $N$) if we only focus on the same model for the same problem. The justification is detailed in~\cref{app:flops}.} the total test-time compute constraint is
\[
\sum_{t=0}^{T-1} \sum_{i=1}^{n_t} k_t^{(i)} \le N.
\]

This framework can express most external TTS control flows: \textit{parallel scaling} maximizes the branching factor in a single step ($T=1$), \textit{sequential scaling} iteratively refines candidates with minimal branching, and \textit{hybrid scaling} alternates between selection, branching, and pruning over multiple rounds. Under this view, disparate discovery algorithms differ primarily in their parameterization of
\[
\bigl(q_t,\; n_t,\; \{k_t^{(i)}\}_{i=1}^{n_t},\; \Pi_t\bigr);
\]
in the common constant-branching case, this reduces to $(q_t, n_t, k_t, \Pi_t)$. Crucially, this abstraction isolates our core research question: assuming access to an informative verifier, how should test-time compute be allocated across selection, branching, and pruning to maximize equation-discovery progress under a fixed budget?

\paragraph{Simplified notation.}
In the remainder of the paper, for simplicity, we focus on time-homogeneous controllers with $n_t \equiv n$, $k_t^{(i)} \equiv k$, and a fixed pruning rule $\Pi_t \equiv \Pi$, and write the controller as $(q, n, k, \Pi)$. We further define the per-iteration search width as $w=nk$.

\section{Method}

\subsection{Diagnosis: Are classical control flows suitable for equation discovery?}

\paragraph{Desiderata: performance and efficiency}
To evaluate the suitability of existing control flows for equation discovery, we establish two primary desiderata: \emph{performance} and \emph{efficiency}. Under a fixed compute budget $N$, \textbf{performance} relies on balancing exploration, which searches broadly over possible functional forms, and exploitation, which refines promising candidates. In our unified framework, this requires a strategic parameterization of the controller; in the simplified setting studied below, this reduces to the selection rule $q$, the number of expanded candidates $n$, the branching factor $k$, and the pruning operator $\Pi$. \textbf{Efficiency}, conversely, is governed by wall-clock time and verification costs. In equation-discovery settings, the verifier may involve fitting parameters, running numerical evaluation, or executing candidate programs. Consequently, higher efficiency directly yields better performance under a fixed time budget. To accelerate wall-clock time, a practical control flow should expose sufficient parallelism by submitting large batches of evaluation jobs simultaneously. When verification is expensive, a useful control flow must carefully tune $(q, n, k, \Pi)$ to push the performance Pareto frontier under a limited evaluation budget.

\begin{figure*}[t]
    \centering
    \includegraphics[width=0.9\textwidth]{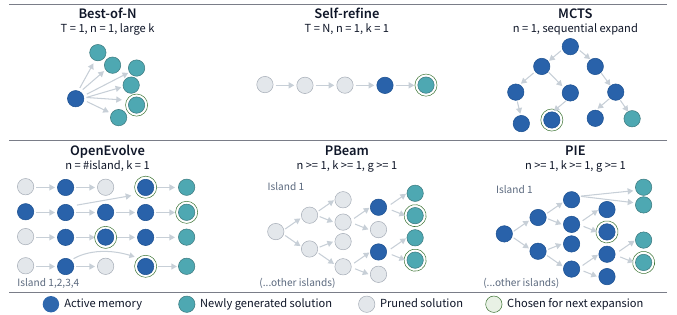}
    \vspace{-0.5em}
    \caption{\textbf{Comparison of TTS control flows under the unified formulation.} Top row: classical baselines emphasize only one aspect of compute allocation: Best-of-$N$ performs purely parallel one-shot sampling ($T=1$, $n=1$, $k\ \text{large}$), Self-refine performs purely sequential single-chain refinement ($T=N$, $n=1$, $k=1$), and MCTS balances exploration and exploitation through sequential tree expansion but with limited parallelism. Bottom row: hybrid, equation-discovery-oriented methods maintain multiple active candidates and combine branching with pruning. OpenEvolve expands multiple isolated islands in parallel ($n=\#\text{islands}$, $k=1$); our Parallel Beam Search (PBeam) performs beam-style expansion within each island ($n\ge1, k\ge1, g\ge1$); and Parallel Iterative Expansion (PIE) further relaxes strict beam pruning by expanding from the full historical pool.
}

    \label{fig:control_flow_comparison}
    \vspace{-1em}
\end{figure*}

\paragraph{Evaluating classical baselines}
How do classical algorithms meet these desiderata in equation discovery? As intuitively illustrated in Figure~\ref{fig:control_flow_comparison}, we evaluate three dominant paradigms through the lens of our $(q, n, k, \Pi)$ abstraction:

\begin{itemize}
   \item \textbf{Extreme Scaling (e.g., Self-Consistency, Self-Refine):} Methods optimizing for pure parallelism ($T=1$, maximal $k$) or pure sequentiality ($n=1$, minimal $k$) fail to balance exploration and exploitation, making them poorly suited for the large search spaces that characterize automated equation discovery.

\item \textbf{Tree Search:} While gracefully balancing exploration and exploitation (e.g., MCTS), sequential node-expansion and backpropagation inherently restrict $n$ and $k$ at any given step. When verifiers are slow, this strict sequentiality creates wall-clock bottlenecks that can reduce overall search efficiency.

\item \textbf{Evolutionary Systems}: Recent self-evolving frameworks (e.g., OpenEvolve~\citep{openevolve}) balance exploration and efficiency by maintaining multiple isolated populations (``islands'') and expanding them sequentially in parallel ($n > 1$). However, these systems are often engineered with complex features such as LLM ensembles, explicit crossover rules, and specialized prompting heuristics. It remains unclear which of these features are necessary for equation discovery.

\end{itemize}

We hypothesize that much of the empirical benefit of evolutionary methods can be explained by a simpler mechanism: improved compute allocation. In our unified $(q, n, k, \Pi)$ view, this suggests a reduced design space centered on a small number of allocation choices: how many candidates are expanded per round, how many continuations each candidate spawns, and how much the search is partitioned into semi-independent groups. This reframing lets us compare heavily engineered systems to minimal controllers and ask directly which allocation decisions matter for performance and efficiency in this setting.

\subsection{Simple yet expressive TTS control flows}

Motivated by this hypothesis, we instantiate a simple family of highly parallel controllers with minimal human-crafted features. We parameterize the search by three practical knobs: population size $n$, per-parent branching factor $k$, and the number of groups $g$. We define the \emph{global width} as $w=nk$, i.e., the total number of expansions produced in one synchronized iteration. The role of $g$ is inspired by island models in evolutionary systems~\citep{novikov2025alphaevolve}, but we treat it more generally as a partitioning variable that changes both search isolation and synchronization granularity. Increasing $g$ creates more independent sub-searches and smaller synchronization units; whether this improves diversity is an empirical question. This reduced $(n,k,g)$ design space is therefore expressive enough to recover useful hybrid control flows while remaining clean enough for systematic study.

To study this reduced design space without confounding domain-specific heuristics, we instantiate two clean, parallel control flows that differ mainly in their selection-and-pruning policies while sharing the same allocation variables $(n,k,g)$:

\textbf{Parallel Beam Search (PBeam):}
We partition the active population and the total budget evenly across $g$ independent groups. Unless otherwise stated, $n$, $k$, and $w=nk$ denote \emph{global} quantities, so each group operates with local population $n/g$, local width $w/g=(n/g)k$, and local budget $N/g$. Omitting the group index for clarity, within each group we execute a constant-width beam search. At step $t$, the selection policy chooses the entire current frontier,
\[
S_t = \mathcal{M}_t,
\qquad
|S_t| = |\mathcal{M}_t| = n/g.
\]
The model generates $k$ candidates per selected candidate, yielding a generated set of size
\[
|G_t| = (n/g)k = w/g.
\]
The pruning operator $\Pi$ strictly retains the top-$n/g$ candidates according to the verifier score $f(x)$:
\[
\mathcal{M}_{t+1} = \operatorname{Top}_{n/g}(\mathcal{M}_t \cup G_t; f),
\]
where $\operatorname{Top}_{m}(A; f)$ denotes the $m$ elements of $A$ with the largest values of $f$.

\textbf{Parallel Iterative Expansion (PIE):}
Building upon PBeam, PIE relaxes the strict frontier-based selection rule within each group to better preserve exploration. Instead of truncating unselected candidates, we maintain the full generation history
\[
\mathcal{H}_{t+1} = \mathcal{H}_t \cup G_t,
\qquad
\mathcal{H}_0 = \mathcal{M}_0.
\]
We then use a softmax selection policy over historical scores to choose $n/g$ candidates for the next expansion step. The probability of selecting a candidate $x \in \mathcal{H}_{t+1}$ is
\[
p_{t+1}(x) = \frac{\exp(f(x)/\tau)}{\sum_{x' \in \mathcal{H}_{t+1}} \exp(f(x')/\tau)},
\]
where $\tau$ is a temperature hyperparameter controlling the exploration--exploitation tradeoff. The pruning operator simply updates the memory state:
\[
\mathcal{M}_{t+1} = \mathcal{H}_{t+1}.
\]
These algorithms instantiate two simple choices of selection-and-pruning policy within the broader $(q_t, n_t, k_t, \Pi_t)$ framework, while keeping the tunable allocation space small and highly parallel. We restrict our current experiments to these two control flows, leaving to future work the optimization of (1) richer selection policies for choosing $S_t$ (e.g., PUCT); (2) adaptive branching methods to dynamically determine $k_t^{(i)}$ for each node $x_t^{(i)}$; (3) diversity-aware pruning operators to maintain population quality during extended rollouts; and (4) iteration-aware allocation for adjusting expansion strategies for different $t$ (e.g., ASHA).

\section{Experiments}

\subsection{Experimental setting}

\paragraph{Benchmarks and models}
We evaluate on LLM-SRBench, a benchmark for scientific equation discovery~\citep{shojaee2025llm}. We focus on the Bio and Material splits, which contain 24 and 25 tasks, respectively. Each task provides a natural-language problem description, variable definitions, a training set, and a held-out test set. Given the problem description and training data, the model proposes a Python program that specifies both the functional form of the equation and an optimizer for fitting its free parameters. During test-time search, the model may observe verifier feedback (fitness) on the training split of previous proposals, but it never accesses the test set. For each random seed, we first average task-level results over all tasks in a domain, and then report the mean and standard deviation over three seeds. Our main study focuses on \texttt{gpt-oss-20b}~\citep{openai2025gptoss} on Bio. We use \texttt{gpt-oss-20b} on Material as a dataset-generalization check, and \texttt{Qwen3-30B-A3B}~\citep{qwen2025qwen3} on Bio as a model-generalization check. We provide more details (e.g., task prompts, dataset statistics, initial solutions) in~\cref{appendix:benchmark} for readers to understand this benchmark.

\paragraph{Search space} We sweep PBeam and PIE under total budgets $N\in\{1,2,4,8,16,32,64,128\}$. We vary the population size $n$, per-node branching factor $k$, group size $g$, and number of iterations $T$, with $N = (n \times k) \times T = w\times T$. Grouping partitions the active population into $g$ independent groups, so the effective width of each group is $w/g$. To keep the sweep manageable, we first fix $g=1$ and sweep $(n,k,T)=(2^i,2^j,2^k)$, where $i+j+k\in[0,7]$ and $i,j,k\in\mathbb{N}$. We then sweep $g$ for the competitive $(n,k,T)$ settings identified in the first stage. Detailed sweep grids, hyperparameters, and results are deferred to~\cref{appendix:experiments}.

\paragraph{Metrics} We follow LLM-SRBench to use $\operatorname{Acc}_{0.1}$ as the primary metric:
\[
\operatorname{Acc}_{0.1}
=
\mathds{1}\!\left(
\max_{1 \le i \le m}
\left|
\frac{\hat{y}_i-y_i}{y_i}
\right|
\le 0.1
\right),
\]
In compute allocation study, we report training $\operatorname{Acc}_{0.1}$ as our primary metric. We defer the interpretation of train--test discrepancies~\cref{sec:discussion}, because our main goal here is to understand compute allocation under a fixed verifier rather than to optimize the verifier itself. We do not report other fitness metrics like NMSE and $R^2$ due to large discrepancy in scales between different tasks, which makes them hard to aggregate.

\subsection{Result analysis}
\label{sec:expr}

\begin{figure*}[t]
    \centering
    \includegraphics[width=\textwidth]{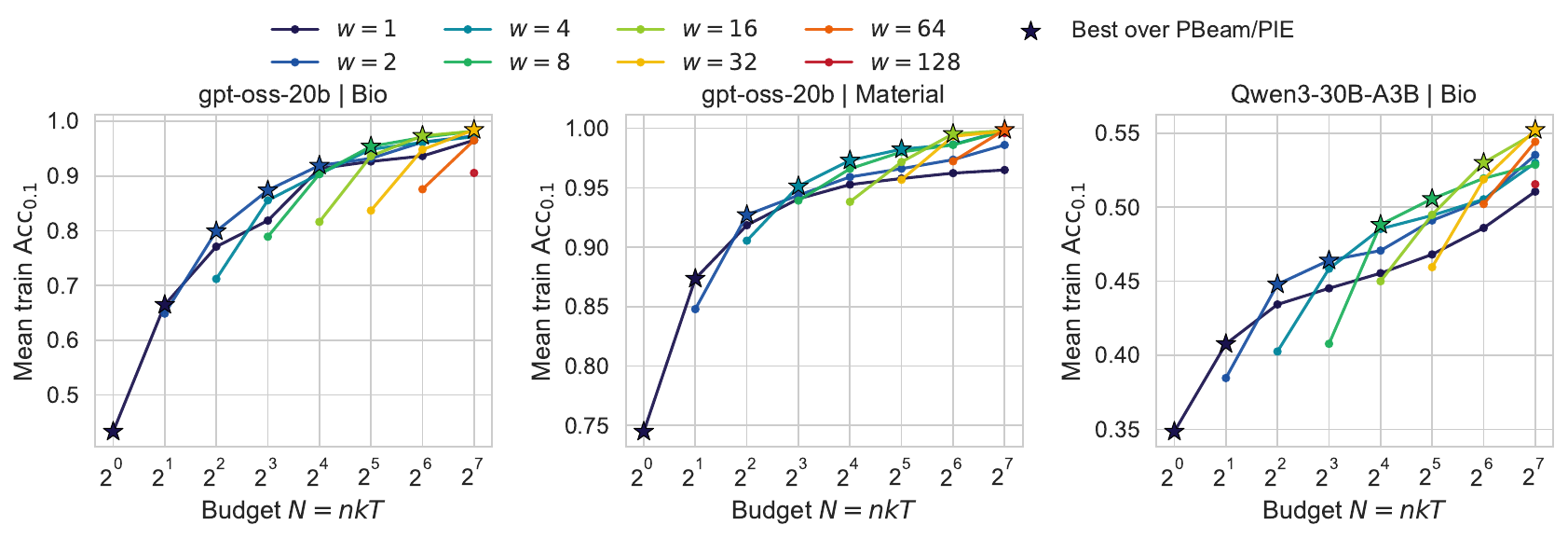}
    \vspace{-2em}
    \caption{
{\textbf{Search-width frontiers across domains and backbones.}
Each curve corresponds to a fixed width $w=nk$ ($g=1$), with colors transitioning from dark (small $w$) to bright (large $w$). Stars mark the best attainable point at each budget over both PBeam and PIE. Across the budgets considered, the envelope tends to shift from narrower to wider searches, indicating that width is a key allocation variable.}}

    \label{fig:width_scaling_main}\vspace{-1em}
\end{figure*}

\paragraph{Search width dictates both performance and wall-clock efficiency.}
We plot the performance envelope over PIE and PBeam for each width and budget in \Cref{fig:width_scaling_main}. Across domains (Bio, Material) and models (\texttt{gpt-oss-20b}, \texttt{Qwen3-30B-A3B}), search width is the most important allocation variable. Very narrow searches over-emphasize sequential refinement, while one-shot wide searches underuse iterative feedback. The best-performing configurations in our sweep usually lie between these extremes.

We deliberately treat the budget--width relationship as an empirical trend rather than a universal scaling law. The sweep contains only a small number of discrete budget levels, and several nearby widths often perform similarly. Nevertheless, the qualitative pattern is consistent: small budgets favor narrow searches, whereas larger budgets tend to benefit from moderately wider synchronized expansion.

\begin{figure*}[t]
    \centering
    \includegraphics[width=\textwidth]{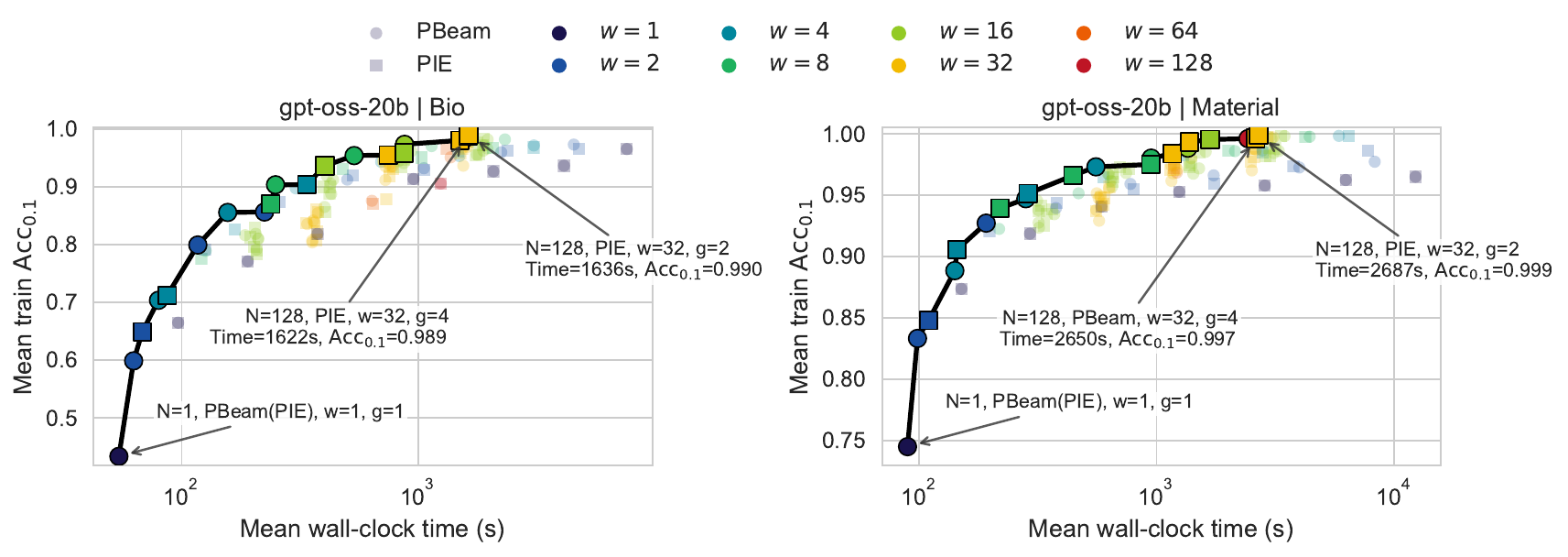}
    \vspace{-2.2em}
    \caption{\textbf{Wall-clock frontier under different compute-allocation strategies.}
    Each point is the best measured configuration at fixed $(N,w,g,\text{algorithm})$ after optimizing over the remaining decomposition. Marker shape indicates the algorithm, color indicates width, and the black curve shows the time--quality Pareto frontier.}
    \label{fig:time_frontier_main}\vspace{-1.2em}
\end{figure*}

Crucially, increasing width can also improve wall-clock efficiency. Larger widths increase model inference parallelism and reduce the number of sequential refinement rounds ($T = N / w$), so the wall-clock Pareto frontier favors moderate-to-large widths (Figure~\ref{fig:time_frontier_main}). For example, on the Bio dataset, shifting from greedy refinement ($w=1$) to width $32$ reduces runtime by nearly $80\%$ (from $7622$\,s to $1586$\,s) while simultaneously improving train $\operatorname{Acc}_{0.1}$ from $0.965$ to $0.984$.

\paragraph{Grouping improves efficiency via reduced synchronization.}
While width dictates the primary time--quality trade-off, grouping acts mainly as a secondary systems knob. Holding the total budget $N$ and the global width $w$ fixed, increasing the number of groups $g$ yields a modest wall-clock speedup because synchronization occurs within groups of size $w/g$ rather than across all $w$ expansions. This reduces idle time from stragglers and improves hardware utilization when model or verifier latency is heterogeneous. Empirically, moving from $g=1$ to $g=8$ yields up to a $1.075\times$ speedup. By contrast, the effect on search quality is weak and not consistently positive: in our equation-discovery benchmark, island-style isolation does not provide a reliable diversity benefit once width is already chosen. Over-grouping can even hurt, because it reduces both the per-group width $w/g$ and the per-group budget $N/g$, pushing each group toward a narrower search regime than the one preferred by the global budget (detailed discussion in \cref{app:wall-clock}). We therefore treat $g$ as a throughput parameter rather than a search-quality parameter.

\paragraph{Other choices are second-order effects.}
Once the global width $w=nk$ is fixed, the remaining design choices have a minor effect on performance. First, the exact decomposition of $w$ into population size $n$ and branching factor $k$ matters much less than choosing the right width in the first place (\cref{tab:nk_assignment}). Second, the difference between PBeam and PIE is also small: their train $\operatorname{Acc}_{0.1}$ gap usually stays within $\pm 2\%$ across budgets and widths (\cref{fig:appendix_algo_delta}). Both effects are substantially smaller than the cross-width differences reported above. In this sense, width determines the operating regime, while the exact $n$--$k$ split and the choice between PBeam and PIE mainly fine-tune performance. We defer the detailed ablations to~\cref{app:second_order}.

\paragraph{An empirical recipe for workflow design.}
Synthesizing these findings, we recommend the following top-down strategy for deploying external test-time search. First, maximize the compute budget by selecting the largest budget $N$ permitted by your inference hardware and latency constraints. Next, determine the search width $w$ using a small pilot sweep over powers of two, or use the empirical width frontier in \Cref{app:scaling_law} as a coarse warm start. Once $w$ has been chosen, set the remaining hyperparameters to moderate values, using balanced choices such as $n/w \ge 1/8$, $k \in [2,8]$, and a mild $g$. Finally, treat the choice among selection rules (e.g., PIE, PBeam, or more complex systems) as a minor refinement step.

\section{Related work}

\paragraph{LLM-driven scientific discovery}
Recent advances in LLMs have enabled them to augment, accelerate, and in some cases automate scientific discovery across the full research pipeline~\citep{ai4science2023impact, wang_scientific_2023}. Scientific discovery is often viewed as two tightly coupled stages: hypothesis generation and empirical verification. Prior work shows that LLMs can act as effective hypothesis generators by drawing on broad domain knowledge and reasoning ability~\citep{qi2023large, si2024can}. When combined with external tools and agentic workflows, they can also execute experiments, analyze data, and iteratively refine candidate hypotheses~\citep{ma2024llm, majumder2024data}. This paradigm has shown promise in domains such as chemistry~\citep{wang2025llm}, biomedicine~\citep{gao_empowering_2024}, and astronomy~\citep{Sun2025Astronomy}. 

\paragraph{Equation discovery}
Equation discovery, or symbolic regression (SR), is a long-standing problem in scientific modeling that seeks equations balancing predictive accuracy and simplicity for a given dataset~\citep{makke2024interpretable}. Classical SR methods mainly rely on search-based optimization~\citep{koza_genetic_1994, virgolin2021improving, cranmer2023interpretable}. Other approaches use Monte Carlo search~\citep{jin2019bayesian, sun2022symbolic} or reinforcement learning~\citep{dsr2019, landajuela2021discovering, mundhenk2021seeding}. More recent work has explored pretrained Transformer-based SR models~\citep{biggio2021neural, valipour2021symbolicgpt, kamienny2022end, vastl2024symformer}. LLM-based SR uses LLMs to generate equation hypotheses and interact with TTS workflows to iteratively improve them while leveraging knowledge encoded in the model~\citep{shojaee2024llm, meyerson2024language, grayeli2024symbolic}.

\section{Extended Discussion}
\label{sec:discussion}

\paragraph{Simple control flows beat complicated systems.}
We also compare the lightweight methods PBeam and PIE with OpenEvolve, as an example of a more engineered evolutionary system. The results show that, when width and depth are controlled to be the same, PBeam and PIE remain competitive. When their parameters are set according to our empirical recipe (\cref{sec:expr}), PBeam and PIE perform slightly better than OpenEvolve, partly because OpenEvolve does not allow $k>1$ or mild grouping with $g<w$, making it less expressive. Detailed slice-level comparisons are reported in \cref{fig:appendix_openevolve_compare}. However, this does not imply that crossover, ensembling, or specialized prompting are never useful. Rather, on this benchmark, their contribution appears secondary to getting the global compute allocation right. An implication: new discovery agents should be compared against width-matched minimal baselines; otherwise, gains from additional engineering may be confounded with gains from simply operating in a better width--depth regime.

\paragraph{Connecting to other TTS literature.}
Our work is complementary to \citet{wu2024inference}, who study compute-optimal inference for closed-ended math reasoning. Under a fixed budget, they ask which combination of model size and existing control flow (tree search vs. sampling) maximizes answer accuracy. In contrast, our setting is open-ended scientific discovery with an external verifier, where the central problem is not only \emph{which} model or \emph{which} control flow to use, but \emph{how} to allocate a fixed budget under a \emph{unified} control flow. 

\paragraph{Verifier quality is the other bottleneck.}
Our conclusions should be read as conditional on the verifier being informative enough to rank candidates meaningfully. In conventional TTS, learned reward models can become the limiting factor because additional search amplifies reward misspecification. Scientific discovery changes the source of the verifier, but not this principle: even when feedback comes from experimental proxies, search can still overfit to the measured objective rather than the true scientific target. In our benchmark, the verifier is based on training-set performance, so the search is optimized against an imperfect proxy (discussed in~\cref{app:train-test}, though the train-test gap is not severe for this benchmark). This suggests a useful decomposition: better control flow improves how efficiently the system explores under a fixed verifier, whereas better verification improves what the system is ultimately driven toward. Both matter, and improvements in verifier fidelity may also shift the optimal exploration--exploitation balance.

\section{Conclusion}

We study test-time scaling for LLM-based equation discovery through the lens of compute allocation. On LLM-SRBench, search width is the dominant factor in our sweep: larger budgets generally favor wider searches, and appropriate width selection improves both accuracy and wall-clock efficiency. These results highlight compute allocation as a central design choice for equation discovery with informative verifiers, and motivate its study in broader open-ended scientific search settings.

% \section*{Author Contributions}

% \section*{Acknowledgments}
% Use unnumbered first level headings for the acknowledgments. All
\section*{Limitations}

This work has several limitations. First, our empirical conclusions are specific to equation discovery as instantiated by LLM-SRBench, especially the Biology and Material Science splits considered in our experiments. Although equation discovery is a useful controlled setting for studying open-ended search with an external verifier, it does not capture the full complexity of scientific discovery, such as experimental design, noisy measurement processes, causal interpretation, safety constraints, or long-horizon laboratory validation. Therefore, the observed width--budget trade-off should be interpreted as evidence for this benchmark setting rather than as a universal law for all scientific-discovery workflows.

Second, our analysis assumes access to an informative verifier. In our experiments, search is guided by training-set performance, and we use training $\operatorname{Acc}_{0.1}$ as the main metric for studying compute allocation. This design helps isolate the effect of test-time compute allocation under a fixed verifier, but it also means that stronger search can overfit to the training split or to the specific proxy objective used by the benchmark. Although the train--test gap is not severe in our reported setting, improved verifier fidelity remains an important bottleneck for applying these methods to real scientific problems.

Third, our study focuses on a limited set of models, domains, and controller families. We evaluate primarily with \texttt{gpt-oss-20b}, include \texttt{Qwen3-30B-A3B} as an auxiliary cross-backbone check, and instantiate two minimal parallel controllers, PBeam and PIE. These choices are sufficient for isolating the role of width, branching, grouping, and depth in a controlled sweep, but they do not cover the full space of possible language models, prompting strategies, learned selection policies, diversity mechanisms, adaptive branching rules, or richer evolutionary systems. As a result, the exact optimal width and controller behavior may change with different backbones, prompts, verifiers, or task families.

Fourth, our compute measure abstracts test-time cost by the number of generated proposals $N$. This is appropriate for comparing allocation strategies under a fixed model and prompt template, but it does not fully capture all practical costs. In real deployments, token lengths, model-serving latency, verifier runtime, hardware utilization, failed executions, and synchronization overhead can vary across tasks and candidate programs. Our wall-clock analysis partially addresses this issue, but a more complete systems-level accounting would be needed before transferring these findings directly to heterogeneous production or laboratory settings.

Finally, our work studies how to allocate inference-time compute more effectively; it does not by itself guarantee scientifically valid discoveries. The generated equations are candidate models optimized against benchmark feedback, and strong benchmark performance should not be interpreted as a substitute for domain expertise, independent validation, or experimental confirmation. This limitation is especially important for high-stakes scientific or engineering applications, where incorrect but plausible equations could lead to misleading conclusions.

\section*{Ethical Considerations}
This work studies test-time compute allocation for scientific equation discovery in a benchmark setting. Our goal is to understand how to allocate inference-time compute more effectively, rather than to automate high-stakes scientific decision-making without oversight.

The main risk of systems of this kind is that they can generate plausible but incorrect equations, hypotheses, or programs, especially when the verifier is imperfect or only measures a proxy objective. In our setting, the search is optimized against benchmark feedback on the training split, so strong search performance should not be interpreted as a validated scientific discovery. Any real-world use would require domain-expert review and additional experimental validation.

A second consideration is computational cost. Large-scale test-time search can consume substantial compute resources. One motivation of this work is therefore efficiency: we study how better compute allocation can improve performance while also reducing wall-clock time and synchronization overhead.

Finally, we plan to release code, prompts, and evaluation scripts for the benchmark setting used in this paper to support reproducibility. However, we do not claim that improved benchmark performance alone implies safe or reliable deployment in real scientific practice.

\bibliography{custom}

@article{kaplan2020scaling,
  title={Scaling laws for neural language models},
  author={Kaplan, Jared and McCandlish, Sam and Henighan, Tom and Brown, Tom B and Chess, Benjamin and Child, Rewon and Gray, Scott and Radford, Alec and Wu, Jeffrey and Amodei, Dario},
  journal={arXiv preprint arXiv:2001.08361},
  year={2020}
}

@software{openevolve,
  title = {OpenEvolve: an open-source evolutionary coding agent},
  author = {Asankhaya Sharma},
  year = {2025},
  publisher = {GitHub},
  url = {https://github.com/algorithmicsuperintelligence/openevolve}
}

@article{gou2023critic,
  title={Critic: Large language models can self-correct with tool-interactive critiquing},
  author={Gou, Zhibin and Shao, Zhihong and Gong, Yeyun and Shen, Yelong and Yang, Yujiu and Duan, Nan and Chen, Weizhu},
  journal={arXiv preprint arXiv:2305.11738},
  year={2023}
}

@inproceedings{chen2024tree,
  title={When is tree search useful for llm planning? it depends on the discriminator},
  author={Chen, Ziru and White, Michael and Mooney, Ray and Payani, Ali and Su, Yu and Sun, Huan},
  booktitle={Proceedings of the 62nd Annual Meeting of the Association for Computational Linguistics (Volume 1: Long Papers)},
  pages={13659--13678},
  year={2024}
}

@article{chen2023teaching,
  title={Teaching large language models to self-debug},
  author={Chen, Xinyun and Lin, Maxwell and Sch{\"a}rli, Nathanael and Zhou, Denny},
  journal={arXiv preprint arXiv:2304.05128},
  year={2023}
}

@article{madaan2023self,
  title={Self-refine: Iterative refinement with self-feedback},
  author={Madaan, Aman and Tandon, Niket and Gupta, Prakhar and Hallinan, Skyler and Gao, Luyu and Wiegreffe, Sarah and Alon, Uri and Dziri, Nouha and Prabhumoye, Shrimai and Yang, Yiming and others},
  journal={Advances in neural information processing systems},
  volume={36},
  pages={46534--46594},
  year={2023}
}

@inproceedings{song2025good,
  title={The good, the bad, and the greedy: Evaluation of llms should not ignore non-determinism},
  author={Song, Yifan and Wang, Guoyin and Li, Sujian and Lin, Bill Yuchen},
  booktitle={Proceedings of the 2025 Conference of the Nations of the Americas Chapter of the Association for Computational Linguistics: Human Language Technologies (Volume 1: Long Papers)},
  pages={4195--4206},
  year={2025}
}

@article{welleck2024meta,
  title   = {From Decoding to Meta-Generation: Inference-time Algorithms for Large Language Models},
  author  = {Sean Welleck and Amanda Bertsch and Matthew Finlayson and Hailey Schoelkopf and Alex Xie and Graham Neubig and Ilia Kulikov and Za{\"i}d Harchaoui},
  journal = {Transactions on Machine Learning Research},
  year    = {2024},
  url     = {https://openreview.net/forum?id=eskQMcIbMS}
}

@article{snell2024scaling,
  title   = {Scaling {LLM} Test-Time Compute Optimally can be More Effective than Scaling Model Parameters},
  author  = {Charlie Snell and Jaehoon Lee and Kelvin Xu and Aviral Kumar},
  journal = {arXiv preprint arXiv:2408.03314},
  year    = {2024},
  doi     = {10.48550/arXiv.2408.03314},
  url     = {https://arxiv.org/abs/2408.03314}
}

@inproceedings{wang2023selfconsistency,
  title     = {Self-Consistency Improves Chain of Thought Reasoning in Language Models},
  author    = {Xuezhi Wang and Jason Wei and Dale Schuurmans and Quoc V. Le and Ed H. Chi and Sharan Narang and Aakanksha Chowdhery and Denny Zhou},
  booktitle = {International Conference on Learning Representations},
  year      = {2023},
  url       = {https://openreview.net/forum?id=1PL1NIMMrw}
}

@article{assumpccao2025codeevolve,
  title={Codeevolve: An open source evolutionary coding agent for algorithm discovery and optimization},
  author={Assump{\c{c}}{\~a}o, Henrique and Ferreira, Diego and Campos, Leandro and Murai, Fabricio},
  journal={arXiv preprint arXiv:2510.14150},
  year={2025}
}

@article{brown2024monkeys,
  title   = {Large Language Monkeys: Scaling Inference Compute with Repeated Sampling},
  author  = {Bradley Brown and Jordan Juravsky and Ryan Ehrlich and Ronald Clark and Quoc V. Le and Christopher R{\'e} and Azalia Mirhoseini},
  journal = {arXiv preprint arXiv:2407.21787},
  year    = {2024},
  doi     = {10.48550/arXiv.2407.21787},
  url     = {https://arxiv.org/abs/2407.21787}
}

@article{cobbe2021trainingverifiers,
  title   = {Training Verifiers to Solve Math Word Problems},
  author  = {Karl Cobbe and Vineet Kosaraju and Mohammad Bavarian and Mark Chen and Heewoo Jun and Lukasz Kaiser and Matthias Plappert and Jerry Tworek and Jacob Hilton and Reiichiro Nakano and Christopher Hesse and John Schulman},
  journal = {arXiv preprint arXiv:2110.14168},
  year    = {2021},
  doi     = {10.48550/arXiv.2110.14168},
  url     = {https://arxiv.org/abs/2110.14168}
}

@inproceedings{yao2023tree,
  title     = {Tree of Thoughts: Deliberate Problem Solving with Large Language Models},
  author    = {Shunyu Yao and Dian Yu and Jeffrey Zhao and Izhak Shafran and Thomas L. Griffiths and Yuan Cao and Karthik Narasimhan},
  booktitle = {Advances in Neural Information Processing Systems},
  volume    = {36},
  year      = {2023},
  url       = {https://proceedings.neurips.cc/paper_files/paper/2023/hash/271db9922b8d1f4dd7aaef84ed5ac703-Abstract-Conference.html}
}

@article{li2025sstar,
  title   = {{S*}: Test Time Scaling for Code Generation},
  author  = {Dacheng Li and Shiyi Cao and Chengkun Cao and Xiuyu Li and Shangyin Tan and Kurt Keutzer and Jiarong Xing and Joseph E. Gonzalez and Ion Stoica},
  journal = {arXiv preprint arXiv:2502.14382},
  year    = {2025},
  doi     = {10.48550/arXiv.2502.14382},
  url     = {https://arxiv.org/abs/2502.14382}
}

@article{lightman2023verify,
  title   = {Let's Verify Step by Step},
  author  = {Hunter Lightman and Vineet Kosaraju and Yura Burda and Harri Edwards and Bowen Baker and Teddy Lee and Jan Leike and John Schulman and Ilya Sutskever and Karl Cobbe},
  journal = {arXiv preprint arXiv:2305.20050},
  year    = {2023},
  doi     = {10.48550/arXiv.2305.20050},
  url     = {https://arxiv.org/abs/2305.20050}
}

@article{romeraparedes2024mathematical,
  title   = {Mathematical Discoveries from Program Search with Large Language Models},
  author  = {Bernardino Romera-Paredes and Mohammadamin Barekatain and Alexander Novikov and Matej Balog and M. Pawan Kumar and Emilien Dupont and Francisco J. R. Ruiz and Jordan S. Ellenberg and Pengming Wang and Omar Fawzi and Pushmeet Kohli and Alhussein Fawzi},
  journal = {Nature},
  volume  = {625},
  number  = {7995},
  pages   = {468--475},
  year    = {2024},
  doi     = {10.1038/s41586-023-06924-6},
  url     = {https://www.nature.com/articles/s41586-023-06924-6}
}

@article{novikov2025alphaevolve,
  title   = {AlphaEvolve: A Coding Agent for Scientific and Algorithmic Discovery},
  author  = {Alexander Novikov and Ng{\^a}n V{\~u} and Marvin Eisenberger and Emilien Dupont and Po-Sen Huang and Adam Zsolt Wagner and Sergey Shirobokov and Borislav Kozlovskii and Francisco J. R. Ruiz and Abbas Mehrabian and M. Pawan Kumar and Abigail See and Swarat Chaudhuri and George Holland and Alex Davies and Sebastian Nowozin and Pushmeet Kohli and Matej Balog},
  journal = {arXiv preprint arXiv:2506.13131},
  year    = {2025},
  doi     = {10.48550/arXiv.2506.13131},
  url     = {https://arxiv.org/abs/2506.13131}
}

@article{lange2025shinkaevolve,
  title={Shinkaevolve: Towards open-ended and sample-efficient program evolution},
  author={Lange, Robert Tjarko and Imajuku, Yuki and Cetin, Edoardo},
  journal={arXiv preprint arXiv:2509.19349},
  year={2025}
}

@article{wang2025thetaevolve,
  title={Thetaevolve: Test-time learning on open problems},
  author={Wang, Yiping and Su, Shao-Rong and Zeng, Zhiyuan and Xu, Eva and Ren, Liliang and Yang, Xinyu and Huang, Zeyi and He, Xuehai and Ma, Luyao and Peng, Baolin and others},
  journal={arXiv preprint arXiv:2511.23473},
  year={2025}
}

@article{yuksekgonul2026learning,
  title={Learning to discover at test time},
  author={Yuksekgonul, Mert and Koceja, Daniel and Li, Xinhao and Bianchi, Federico and McCaleb, Jed and Wang, Xiaolong and Kautz, Jan and Choi, Yejin and Zou, James and Guestrin, Carlos and others},
  journal={arXiv preprint arXiv:2601.16175},
  year={2026}
}

@article{shojaee2025llm,
  title   = {{LLM-SRBench}: A New Benchmark for Scientific Equation Discovery with Large Language Models},
  author  = {Parshin Shojaee and Ngoc-Hieu Nguyen and Kazem Meidani and Amir Barati Farimani and Khoa D. Doan and Chandan K. Reddy},
  journal = {arXiv preprint arXiv:2504.10415},
  year    = {2025},
  doi     = {10.48550/arXiv.2504.10415},
  url     = {https://arxiv.org/abs/2504.10415}
}

@misc{openai2025gptoss,
  title        = {gpt-oss-120b \& gpt-oss-20b Model Card},
  author       = {{OpenAI}},
  year         = {2025},
  month        = aug,
  howpublished = {\url{https://openai.com/index/gpt-oss-model-card/}},
  note         = {Published August 5, 2025}
}

@article{qwen2025qwen3,
  title   = {Qwen3 Technical Report},
  author  = {An Yang and Anfeng Li and Baosong Yang and Beichen Zhang and Binyuan Hui and Bo Zheng and Bowen Yu and Chang Gao and Chengen Huang and Chenxu Lv and Chujie Zheng and Dayiheng Liu and Fan Zhou and Fei Huang and Feng Hu and Hao Ge and Haoran Wei and Huan Lin and Jialong Tang and Jian Yang and Jianhong Tu and Jianwei Zhang and others},
  journal = {arXiv preprint arXiv:2505.09388},
  year    = {2025},
  doi     = {10.48550/arXiv.2505.09388},
  url     = {https://arxiv.org/abs/2505.09388}
}

@article{wu2024inference,
  title={Inference scaling laws: An empirical analysis of compute-optimal inference for problem-solving with language models},
  author={Wu, Yangzhen and Sun, Zhiqing and Li, Shanda and Welleck, Sean and Yang, Yiming},
  journal={arXiv preprint arXiv:2408.00724},
  year={2024}
}

@article{guan2025rstar,
  title={RStar-math: Small LLMs can master math reasoning with self-evolved deep thinking},
  author={Guan, Xinyu and Zhang, Li Lyna and Liu, Yifei and Shang, Ning and Sun, Youran and Zhu, Yi and Yang, Fan and Yang, Mao},
  journal={arXiv preprint arXiv:2501.04519},
  year={2025}
}

@article{wang_scientific_2023,
	title = {Scientific discovery in the age of artificial intelligence},
	volume = {620},
	issn = {1476-4687},
	url = {https://doi.org/10.1038/s41586-023-06221-2},
	doi = {10.1038/s41586-023-06221-2},
	number = {7972},
	journal = {Nature},
	author = {Wang, Hanchen and Fu, Tianfan and Du, Yuanqi and Gao, Wenhao and Huang, Kexin and Liu, Ziming and Chandak, Payal and Liu, Shengchao and Van Katwyk, Peter and Deac, Andreea and Anandkumar, Anima and Bergen, Karianne and Gomes, Carla P. and Ho, Shirley and Kohli, Pushmeet and Lasenby, Joan and Leskovec, Jure and Liu, Tie-Yan and Manrai, Arjun and Marks, Debora and Ramsundar, Bharath and Song, Le and Sun, Jimeng and Tang, Jian and Veličković, Petar and Welling, Max and Zhang, Linfeng and Coley, Connor W. and Bengio, Yoshua and Zitnik, Marinka},
	month = aug,
	year = {2023},
	pages = {47--60},
}

@article{ai4science2023impact,
  title={The impact of large language models on scientific discovery: a preliminary study using gpt-4},
  author={AI4Science, Microsoft Research and Quantum, Microsoft Azure},
  journal={arXiv preprint arXiv:2311.07361},
  year={2023}
}

@ARTICLE{Sun2025Astronomy,
       author = {{Sun}, Zechang and {Ting}, Yuan-Sen and {Liang}, Yaobo and {Duan}, Nan and {Huang}, Song and {Cai}, Zheng},
        title = "{Mephisto: Self-Improving Large Language Model-Based Agents for Automated Interpretation of Multi-band Galaxy Observations}",
      journal = {arXiv e-prints},
         year = 2025,
        month = oct,
          eid = {arXiv:2510.08354},
        pages = {arXiv:2510.08354},
          doi = {10.48550/arXiv.2510.08354},
archivePrefix = {arXiv},
       eprint = {2510.08354},
 primaryClass = {astro-ph.IM},
       adsurl = {https://ui.adsabs.harvard.edu/abs/2025arXiv251008354S}
}

@article{wang2025llm,
  title={LLM-Augmented Chemical Synthesis and Design Decision Programs},
  author={Haorui Wang and Jeff Guo and Lingkai Kong and Rampi Ramprasad and Philippe Schwaller and Yuanqi Du and Chao Zhang},
  journal={Forty-Second International Conference on Machine Learning},
  year={2025},
  url={https://openreview.net/forum?id=NhkNX8jYld}
}

@article{qi2023large,
  title={Large language models are zero shot hypothesis proposers},
  author={Qi, Biqing and Zhang, Kaiyan and Li, Haoxiang and Tian, Kai and Zeng, Sihang and Chen, Zhang-Ren and Zhou, Bowen},
  journal={arXiv preprint arXiv:2311.05965},
  year={2023}
}

@article{si2024can,
  title={Can llms generate novel research ideas? a large-scale human study with 100+ nlp researchers},
  author={Si, Chenglei and Yang, Diyi and Hashimoto, Tatsunori},
  journal={arXiv preprint arXiv:2409.04109},
  year={2024}
}

@article{ma2024llm,
  title={Llm and simulation as bilevel optimizers: A new paradigm to advance physical scientific discovery},
  author={Ma, Pingchuan and Wang, Tsun-Hsuan and Guo, Minghao and Sun, Zhiqing and Tenenbaum, Joshua B and Rus, Daniela and Gan, Chuang and Matusik, Wojciech},
  journal={arXiv preprint arXiv:2405.09783},
  year={2024}
}

@article{majumder2024data,
  title={Data-driven discovery with large generative models},
  author={Majumder, Bodhisattwa Prasad and Surana, Harshit and Agarwal, Dhruv and Hazra, Sanchaita and Sabharwal, Ashish and Clark, Peter},
  journal={arXiv preprint arXiv:2402.13610},
  year={2024}
}

@article{shojaee2024llm,
  title={Llm-sr: Scientific equation discovery via programming with large language models},
  author={Shojaee, Parshin and Meidani, Kazem and Gupta, Shashank and Farimani, Amir Barati and Reddy, Chandan K},
  journal={arXiv preprint arXiv:2404.18400},
  year={2024}
}

@article{gao_empowering_2024,
	title = {Empowering biomedical discovery with {AI} agents},
	volume = {187},
	issn = {0092-8674},
	url = {https://doi.org/10.1016/j.cell.2024.09.022},
	doi = {10.1016/j.cell.2024.09.022},
	number = {22},
	urldate = {2026-03-28},
	journal = {Cell},
	publisher = {Elsevier},
	author = {Gao, Shanghua and Fang, Ada and Huang, Yepeng and Giunchiglia, Valentina and Noori, Ayush and Schwarz, Jonathan Richard and Ektefaie, Yasha and Kondic, Jovana and Zitnik, Marinka},
	month = oct,
	year = {2024},
	pages = {6125--6151},
}

@article{koza_genetic_1994,
	title = {Genetic programming as a means for programming computers by natural selection},
	volume = {4},
	issn = {1573-1375},
	url = {https://doi.org/10.1007/BF00175355},
	doi = {10.1007/BF00175355},
	number = {2},
	journal = {Statistics and Computing},
	author = {Koza, John R.},
	month = jun,
	year = {1994},
	pages = {87--112},
}

@article{cranmer2023interpretable,
  title={Interpretable machine learning for science with PySR and SymbolicRegression. jl},
  author={Cranmer, Miles},
  journal={arXiv preprint arXiv:2305.01582},
  year={2023}
}

@article{virgolin2021improving,
  title={Improving model-based genetic programming for symbolic regression of small expressions},
  author={Virgolin, Marco and Alderliesten, Tanja and Witteveen, Cees and Bosman, Peter AN},
  journal={Evolutionary computation},
  volume={29},
  number={2},
  pages={211--237},
  year={2021},
  publisher={MIT Press One Rogers Street, Cambridge, MA 02142-1209, USA journals-info~…}
}

@article{makke2024interpretable,
  title={Interpretable scientific discovery with symbolic regression: a review},
  author={Makke, Nour and Chawla, Sanjay},
  journal={Artificial Intelligence Review},
  volume={57},
  number={1},
  pages={2},
  year={2024},
  publisher={Springer}
}

@article{sun2022symbolic,
  title={Symbolic physics learner: Discovering governing equations via monte carlo tree search},
  author={Sun, Fangzheng and Liu, Yang and Wang, Jian-Xun and Sun, Hao},
  journal={arXiv preprint arXiv:2205.13134},
  year={2022}
}

@article{jin2019bayesian,
  title={Bayesian symbolic regression},
  author={Jin, Ying and Fu, Weilin and Kang, Jian and Guo, Jiadong and Guo, Jian},
  journal={arXiv preprint arXiv:1910.08892},
  year={2019}
}

@article{dsr2019,
  title={Deep symbolic regression: Recovering mathematical expressions from data via risk-seeking policy gradients},
  author={Petersen, Brenden K and Landajuela, Mikel and Mundhenk, T Nathan and Santiago, Claudio P and Kim, Soo K and Kim, Joanne T},
  journal={arXiv preprint arXiv:1912.04871},
  year={2019}
}

@inproceedings{landajuela2021discovering,
  title={Discovering symbolic policies with deep reinforcement learning},
  author={Landajuela, Mikel and Petersen, Brenden K and Kim, Sookyung and Santiago, Claudio P and Glatt, Ruben and Mundhenk, Nathan and Pettit, Jacob F and Faissol, Daniel},
  booktitle={International Conference on Machine Learning},
  pages={5979--5989},
  year={2021},
  organization={PMLR}
}

@inproceedings{mundhenk2021seeding,
  title={Symbolic Regression via Neural-Guided Genetic Programming Population Seeding},
  author={T. Nathan Mundhenk and Mikel Landajuela and Ruben Glatt and Claudio P. Santiago and Daniel M. Faissol and Brenden K. Petersen},
  booktitle={Advances in Neural Information Processing Systems},
  year={2021}
}

@inproceedings{biggio2021neural,
  title={Neural symbolic regression that scales},
  author={Biggio, Luca and Bendinelli, Tommaso and Neitz, Alexander and Lucchi, Aurelien and Parascandolo, Giambattista},
  booktitle={International conference on machine learning},
  pages={936--945},
  year={2021},
  organization={Pmlr}
}

@article{valipour2021symbolicgpt,
  title={Symbolicgpt: A generative transformer model for symbolic regression},
  author={Valipour, Mojtaba and You, Bowen and Panju, Maysum and Ghodsi, Ali},
  journal={arXiv preprint arXiv:2106.14131},
  year={2021}
}

@article{kamienny2022end,
  title={End-to-end symbolic regression with transformers},
  author={Kamienny, Pierre-Alexandre and d'Ascoli, St{\'e}phane and Lample, Guillaume and Charton, Fran{\c{c}}ois},
  journal={Advances in Neural Information Processing Systems},
  volume={35},
  pages={10269--10281},
  year={2022}
}

@article{vastl2024symformer,
  title={Symformer: End-to-end symbolic regression using transformer-based architecture},
  author={Vastl, Martin and Kulh{\'a}nek, Jon{\'a}{\v{s}} and Kubal{\'\i}k, Ji{\v{r}}{\'\i} and Derner, Erik and Babu{\v{s}}ka, Robert},
  journal={IEEE Access},
  volume={12},
  pages={37840--37849},
  year={2024},
  publisher={IEEE}
}

@article{meyerson2024language,
  title={Language model crossover: Variation through few-shot prompting},
  author={Meyerson, Elliot and Nelson, Mark J and Bradley, Herbie and Gaier, Adam and Moradi, Arash and Hoover, Amy K and Lehman, Joel},
  journal={ACM Transactions on Evolutionary Learning},
  volume={4},
  number={4},
  pages={1--40},
  year={2024},
  publisher={ACM New York, NY}
}

@article{grayeli2024symbolic,
  title={Symbolic regression with a learned concept library},
  author={Grayeli, Arya and Sehgal, Atharva and Costilla-Reyes, Omar and Cranmer, Miles and Chaudhuri, Swarat},
  journal={Advances in Neural Information Processing Systems},
  volume={37},
  pages={44678--44709},
  year={2024}
}

\newpage

\appendix

\section{Compute budget $N$ and FLOPs}
\label{app:flops}

Our unified formulation measures test-time compute by the total number of generation passes,
\[
N \;=\; \sum_{t=0}^{T-1}\sum_{i=1}^{n_t} k_t^{(i)},
\]
namely, the total number of candidate proposals produced across all steps. While prior test-time scaling work often reports inference compute in FLOPs, under our experimental setting this choice is equivalent to measuring compute by $N$.

To see this, consider a single generation pass that produces one proposal. Let its input and output lengths be $L_{\mathrm{in}}$ and $L_{\mathrm{out}}$, respectively. Following standard scaling-law analyses~\citep{kaplan2020scaling}, the FLOPs required for one autoregressive rollout of a transformer with $P$ parameters is approximately
\[
C_{\mathrm{rollout}}
\;\approx\;
2P\bigl(L_{\mathrm{in}} + L_{\mathrm{out}}\bigr).
\]
Now index all generated proposals in the entire search process by $(t,i,j)$, where $t$ is the iteration, $i$ indexes the selected parent candidate, and $j \in \{1,\dots,k_t^{(i)}\}$ indexes the newly generated proposal. The total inference cost is then
\[
C_{\mathrm{total}}
\;\approx\;
\sum_{t=0}^{T-1}\sum_{i=1}^{n_t}\sum_{j=1}^{k_t^{(i)}}
2P\bigl(L_{\mathrm{in}}^{(t,i,j)} + L_{\mathrm{out}}^{(t,i,j)}\bigr).
\]

Under our controlled setting, the base model is fixed, so $P$ is constant. Moreover, we study compute allocation for the same task instance: each rollout takes as context a previous program (or equation) and asks the model to produce a refined or mutated proposal of comparable size. Hence both the input and output lengths stay within a task-dependent bounded range. That is, there exist constants
\[
\underline{L}_{\mathrm{in}},\ \overline{L}_{\mathrm{in}},\ 
\underline{L}_{\mathrm{out}},\ \overline{L}_{\mathrm{out}}
\]
such that for every rollout,
\[
\underline{L}_{\mathrm{in}}
\le
L_{\mathrm{in}}^{(t,i,j)}
\le
\overline{L}_{\mathrm{in}},
\qquad
\underline{L}_{\mathrm{out}}
\le
L_{\mathrm{out}}^{(t,i,j)}
\le
\overline{L}_{\mathrm{out}}.
\]
Therefore, each rollout has cost bounded between two constants:
\[
c_{\min}
\;\le\;
C_{\mathrm{rollout}}^{(t,i,j)}
\;\le\;
c_{\max},
\]
where
\[
c_{\min}=2P\bigl(\underline{L}_{\mathrm{in}}+\underline{L}_{\mathrm{out}}\bigr),
\qquad
c_{\max}=2P\bigl(\overline{L}_{\mathrm{in}}+\overline{L}_{\mathrm{out}}\bigr).
\]
Summing over all rollouts gives
\[
c_{\min} N
\;\le\;
C_{\mathrm{total}}
\;\le\;
c_{\max} N.
\]
Hence, for a fixed model on a fixed task, the total test-time FLOPs grow linearly with $N$. In practice, $c_{\min}$ is also close to $c_{\max}$ as each input and output consists of the same prompt template and a solution. In other words, $N$ and FLOPs are equivalent up to a task-dependent constant factor, and therefore induce the same notion of compute budget to compare different compute-allocation strategies.

Finally, our budget $N$ accounts only for the LLM generation cost. Auxiliary operations such as selection, pruning, and verifier evaluation are either negligible relative to model inference or held fixed across methods and thus do not affect the equivalence above.

\section{Benchmark and implementation details}
\label{appendix:benchmark}

\subsection{Benchmark instantiation}

Our experiments instantiate the LSR-Synth portion of LLM-SRBench and restrict attention to the two domains studied in the main paper: Biology and Material Science. At the benchmark level, these correspond to 24 biology tasks and 25 material-science tasks, yielding 49 tasks in total. In our local benchmark layout, each task directory contains four components: (i) a natural-language task specification, (ii) a shared executable scaffold, (iii) a task-specific evaluator, and (iv) a local dataset partitioned into \texttt{train}, \texttt{test}, and \texttt{ood\_test} splits. The search procedure is given access only to the task specification and the training split. The held-out \texttt{test} split is queried exclusively for logging best-so-far in-domain generalization, while \texttt{ood\_test} is never used for model selection.

Following the benchmark construction protocol, each synthetic task is generated from 5{,}000 samples. In the task files used by our implementation, 4{,}000 samples are exposed through the training split and the remainder are reserved for held-out test evaluation. Across domains, Biology tasks share the mapping $(t,P)\mapsto dP/dt$, whereas Material tasks share $(\epsilon,T)\mapsto \sigma$; task-level variation arises entirely from the hidden symbolic law and from the associated numeric ranges encoded in the benchmark files.

\begin{table*}[H]
    \centering
    \setlength{\tabcolsep}{4pt}
    \caption{\textbf{Dataset statistics for the two LSR-Synth domains considered in the paper.} The input/output signatures are domain-level invariants; the hidden equation and numeric ranges vary across tasks.}
    \small
    \begin{tabularx}{\linewidth}{@{}l c c c c X@{}}
        \toprule
        Domain & \# Tasks & Scientific setting & Inputs & Output & Range summary from task files \\
        \midrule
        Bio & 24 & Population dynamics & $(t, P)$ & $dP/dt$ & Each task exposes 4{,}000 training points. Across tasks, $t$ lies near $[1, 54.10]$, the upper bound of $P$ ranges from $1$ to $996.63$, and the training output range for $dP/dt$ spans approximately $[-33.35, 60.54]$. \\
        Material & 25 & Stress--strain relationships & $(\epsilon, T)$ & $\sigma$ & Each task exposes 4{,}000 training points. Across tasks, $\epsilon$ lies near $[0, 0.54]$, $T$ lies near $[273, 542.99]\,\mathrm{K}$, and the training output range for $\sigma$ spans approximately $[-606.19, 1.36\times 10^5]$. \\
        \bottomrule
    \end{tabularx}

    \label{tab:appendix_benchmark_stats}
\end{table*}

\subsection{Prompt construction and executable scaffold}

Our prompting interface mirrors the executable structure of the benchmark while remaining intentionally minimal. Each prompt is assembled from three pieces: the task instruction, a fixed program scaffold split into \texttt{EXACT\_PREFIX} and \texttt{EXACT\_SUFFIX} around the editable region, and at most one in-context inspiration drawn from the current best valid program in the same group. The implementation hard-caps the number of inspirations at one. At decoding time, only the code between \texttt{\# EVOLVE-BLOCK-START} and \texttt{\# EVOLVE-BLOCK-END} is extracted from the model response; the full runnable program is then reconstructed automatically from the fixed prefix, the generated block, and the fixed suffix. To keep the search space comparable across runs, the only permitted third-party libraries are \texttt{numpy}, \texttt{scipy}, and \texttt{scikit-learn}.

\paragraph{Template sketch.}
\lstset{
    basicstyle=\ttfamily\small, % 使用小一号的字号，或用 \footnotesize
    breaklines=true,            % 强制开启自动换行
    breakatwhitespace=false,    % 允许在非空格处换行，防止单词过长冲出边界
    keepspaces=true,            % 保持空格
    columns=flexible            % 使用弹性的列宽，可以让字符排列得更紧凑
}

\begin{lstlisting}
Task: <task instruction>

Improve the solution:

Score: <score>
Metrics:
  acc01: <value>
  combined_score: <value>
  ...

Code:
```python
<best prior program in the current group>
```

Generation instruction (must follow exactly):
1) Only the code between # EVOLVE-BLOCK-START and # EVOLVE-BLOCK-END is extracted.
2) The final program is reconstructed as EXACT_PREFIX + evolved_block + EXACT_SUFFIX.
3) Keep marker lines exactly as written.
4) Return one Python code block that includes both EVOLVE-BLOCK markers.

EXACT_PREFIX (kept unchanged):
```python
<prefix ending with # EVOLVE-BLOCK-START>
```

EXACT_SUFFIX (kept unchanged):
```python
<suffix starting from # EVOLVE-BLOCK-END>
```

Available packages: numpy, scipy, scikit-learn
\end{lstlisting}
\paragraph{Shared initialization.}
All 49 Biology and Material tasks use the same initialization scaffold. The initializer is deliberately weak: it fits an affine baseline to the raw inputs by least squares and therefore serves as a domain-agnostic starting point shared across all runs and all groups.

\begin{lstlisting}
# EVOLVE-BLOCK-START
"""Initial program: affine symbolic-regression baseline fit by least squares."""

import numpy as np


def scaling_law_func(data_points, params):
    X = np.atleast_2d(np.asarray(data_points, dtype=float))
    params = np.asarray(params, dtype=float)

    if params.ndim == 1:
        params = params[None, :]

    n_features = X.shape[1]
    expected = n_features + 1
    if params.shape[1] < expected:
        pad = np.zeros((params.shape[0], expected - params.shape[1]), dtype=float)
        params = np.concatenate([params, pad], axis=1)

    weights = params[:, :n_features]
    bias = params[:, n_features]

    pred = X @ weights.T + bias[None, :]
    return pred[:, 0] if pred.shape[1] == 1 else pred


def fit_scaling_law(data_points, loss_values):
    X = np.atleast_2d(np.asarray(data_points, dtype=float))
    y = np.asarray(loss_values, dtype=float)

    if y.ndim == 1:
        y = y[:, None]

    design = np.concatenate([X, np.ones((X.shape[0], 1), dtype=float)], axis=1)
    coeffs, _, _, _ = np.linalg.lstsq(design, y, rcond=None)

    params = coeffs.T
    return params[0] if params.shape[0] == 1 else params
# EVOLVE-BLOCK-END
\end{lstlisting}

\subsection{Grouped TTS search and implementation details}

Algorithm~\ref{alg:appendix_tts_impl} summarizes the grouped tree-structured search used throughout our experiments. Relative to the notation in the main text, the implementation uses $(n,k,g)$ for population size, branching factor, and number of groups, respectively. The synchronized search depth is
\[
T=\max\!\left(1,\left\lfloor \frac{N}{nk}\right\rfloor\right),
\]
and the total width expanded per synchronized iteration is $w=nk$.

\begin{algorithm*}[h]
    \caption{Grouped TTS search used in our experiments}
    \label{alg:appendix_tts_impl}
    \begin{algorithmic}[1]
        \Require problem $p$, budget $N$, population size $n$, branching factor $k$, groups $g$, algorithm $\in \{\text{PBeam}, \text{PIE}\}$
        \State $T \gets \max(1,\lfloor N/(nk)\rfloor)$, $s \gets n/g$
        \State evaluate the shared initial program on the training split to obtain node $x_0$
        \State initialize memory $\mathcal{M}_0$ by replicating $x_0$ into each of the $g$ groups with $s$ slots per group
        \For{$t = 1$ to $T$}
            \State sample $s$ parents per group
            \Statex \hspace{1.5em}\text{PBeam}: choose the top-$s$ valid nodes by train $\operatorname{Acc}_{0.1}$
            \Statex \hspace{1.5em}\text{PIE}: sample without replacement from a softmax over clipped scores
            \ForAll{selected parent $x$}
                \For{$j = 1$ to $k$}
                    \State build a prompt from the task instruction, the fixed scaffold, and the best valid group-local inspiration
                    \State query the LLM and extract only the code inside the \texttt{EVOLVE-BLOCK}
                    \State evaluate the reconstructed program on the training split; set its search score to train $\operatorname{Acc}_{0.1}$
                \EndFor
            \EndFor
            \If{algorithm $=$ \text{PBeam}}
                \State in each group, keep the top-$s$ nodes from $\mathcal{M}_{t-1} \cup \{\text{new children}\}$
            \Else
                \State append children to memory and cap each group to $\max(4s, N)$ nodes
            \EndIf
            \State evaluate the current best-so-far node on the held-out \texttt{test} split for logging only
        \EndFor
        \State \Return best-scoring node seen during the run
    \end{algorithmic}
\end{algorithm*}

Although the evaluator also reports NMSE, $R^2$, and a benchmark-specific \texttt{combined\_score}, the search objective is always training $\operatorname{Acc}_{0.1}$. In the released PIE implementation, parent selection is parameter free: scores are clipped to $[-10,10]$ for numerical stability and sampled at unit temperature. All generations and evaluator calls are dispatched asynchronously under global concurrency caps, and every evaluation is executed in a fresh Python subprocess with a 600-second timeout.

\subsection{Sweep grid and runtime settings}
\label{appendix:experiments}

\begin{table*}[h]
    \centering
    \small
    \setlength{\tabcolsep}{4pt}
    \caption{\textbf{Experimental grid used in the reported study.}}
    \begin{tabularx}{\linewidth}{@{}l X@{}}
        \toprule
        Study stage & Configurations \\
        \midrule
        $nk$ assignment & Main width--depth sweep. For each reported model/domain/seed tuple, we fix $g=1$ and enumerate all power-of-two pairs $(n,k)$ satisfying $nk \le 128$. This yields 36 \textsc{PBeam} configurations. \textsc{PIE} uses the same grid except for the degenerate $(n,k)=(1,1)$ case and the single-step settings with $nk=128$, yielding 27 additional configurations. Each run is launched once at maximum budget $128$, and lower target budgets $N\in\{1,2,4,8,16,32,64,128\}$ are recovered from best-so-far checkpoints by selecting the last logged state with cumulative \texttt{budget\_used} $\le N$. \\
        G assignment & Grouping ablation, performed only for \texttt{gpt-oss-20b} on Bio and Material. We fix $N=128$ and sweep group counts on the competitive high-width slices identified in the first stage. Width-32 settings are $(n,k)\in\{(32,1),(16,2),(8,4)\}$ and width-16 settings are $(n,k)\in\{(16,1),(8,2)\}$. For each pair we test every divisor $g$ of $n$ in $\{1,2,4,8,16,32\}$, producing 24 configurations per algorithm. \\
        \bottomrule
    \end{tabularx}
    \label{tab:appendix_grid}
\end{table*}

Unless otherwise stated, all execution uses temperature $0.7$, \texttt{max\_tokens}=16384, three random seeds $\{0,1,2\}$, and global concurrency caps of 128 LLM calls and 128 evaluator subprocesses. The main paper reports \texttt{gpt-oss-20b} on Bio and Material, and \texttt{Qwen3-30B-A3B-Thinking-2507} on Bio as a cross-backbone check. Grouping ablations are run only for \texttt{gpt-oss-20b}. Held-out \texttt{test} performance is measured after every synchronized iteration, but it is never fed back into the search loop.

\subsection{Ground-truth symbolic laws for the studied domains}

For completeness, we reproduce the benchmark ground-truth equations for all 49 LSR-Synth tasks considered in this paper. The task identifiers follow the naming convention in LLM-SRBench Table~4.

\begin{table*}[htbp]
\centering
\small % 保持较小的字号
\newcolumntype{L}[1]{>{\raggedright\arraybackslash}p{#1}} % 保持原有的靠左对齐并支持换行定义

\setlength{\tabcolsep}{3pt}       % 保持原有的列间距微调
\renewcommand{\arraystretch}{1.2} % 稍微调大一点点行高（1.12调到1.2），让含有分式和指数的复杂公式上下更有呼吸感

\caption{\textbf{Biology-domain LSR-Synth equations used in this paper.} Each entry reports the right-hand side of $\frac{dP}{dt}=f(t,P)$.}
\label{tab:appendix_bio_equations}

% 使用普通的 tabular，总宽度比例：0.08 + 0.39 + 0.08 + 0.39 = 0.94\textwidth，给右侧留出了一点安全边缘防止溢出
\begin{tabular}{@{} L{0.08\textwidth} L{0.39\textwidth} L{0.08\textwidth} L{0.39\textwidth} @{}}
\toprule
ID & Right-hand side & ID & Right-hand side \\
\midrule

BPG1  & $r\left(1-\frac{P(t)}{K_0}\right)P(t)+rP(t)^{0.33}$  
      & BPG13 & $\beta P(t)\sin(\omega t)+r\left(1-\frac{P(t)}{K_0}\right)P(t)$ \\ \addlinespace

BPG2  & $rP(t)\exp(-\gamma t)+\frac{rP(t)^2}{\alpha P(t)+1}$  
      & BPG14 & $r\left(-1+\frac{P(t)}{\alpha}\right)\left(1-\frac{P(t)}{K_0}\right)P(t)+rP(t)+\frac{rP(t)}{1+\exp\left(-\alpha(-\beta+P(t))\right)}$ \\ \addlinespace

BPG3  & $\beta P(t)\sin(\omega t)+rP(t)\exp(-\gamma t)$  
      & BPG15 & $r\left(1-\frac{P(t)}{K_0}\right)P(t)+r\left(1-\exp(-\gamma P(t))\right)P(t)+rP(t)\exp(-\gamma t)$ \\ \addlinespace

BPG4  & $r\left(-1+\frac{P(t)}{\alpha}\right)\left(1-\frac{P(t)}{K_0}\right)P(t)+r\left(1-\exp(-\gamma P(t))\right)P(t)$ 
      & BPG16 & $rP(t)^{0.33}+rP(t)\exp(-\gamma t)$ \\ \addlinespace

BPG5  & $r\left(1-\frac{P(t)}{K_0}\right)P(t)+\frac{rP(t)}{1+\exp\left(-\alpha(-\beta+P(t))\right)}$ 
      & BPG17 & $r\left(-1+\frac{P(t)}{\alpha}\right)\left(1-\frac{P(t)}{K_0}\right)P(t)+rP(t)^{0.33}+rP(t)$ \\ \addlinespace

BPG6  & $r\left(1-\frac{P(t)}{K_0}\right)P(t)+\frac{rP(t)^2}{\alpha P(t)+1}$  
      & BPG18 & $r\left(-1+\frac{P(t)}{\alpha}\right)\left(1-\frac{P(t)}{K_0}\right)P(t)+rP(t)^{0.33}$ \\ \addlinespace

BPG7  & $-Q\alpha P(t)+r\left(1-\frac{P(t)}{K_0}\right)P(t)+rP(t)^{0.33}+rP(t)$ 
      & BPG19 & $\beta P(t)\sin(\omega t)+r\left(1-\frac{P(t)}{K_0}\right)P(t)+rP(t)$ \\ \addlinespace

BPG8  & $r\left(-1+\frac{P(t)}{\alpha}\right)\left(1-\frac{P(t)}{K_0}\right)P(t)+r\left(1-\frac{P(t)}{K_0}\right)P(t)+rP(t)^{0.33}$ 
      & BPG20 & $r\left(1-\frac{P(t)}{K_0}\right)P(t)+\frac{rP(t)}{t^{\alpha}}$ \\ \addlinespace

BPG9  & $r\left(1-\frac{P(t)}{K_0}\right)P(t)+rP(t)^{0.33}+rP(t)$ 
      & BPG21 & $r\left(-1+\frac{P(t)}{\alpha}\right)\left(1-\frac{P(t)}{K_0}\right)P(t)+r\left(1-\frac{P(t)}{K_0}\right)P(t)+\frac{rP(t)}{1+\exp\left(-\alpha(-\beta+P(t))\right)}$ \\ \addlinespace

BPG10 & $r\left(-1+\frac{P(t)}{\alpha}\right)\left(1-\frac{P(t)}{K_0}\right)P(t)+r\left(1-\frac{P(t)}{K_0}\right)P(t)+r\left(1-\exp(-\gamma P(t))\right)P(t)$ 
      & BPG22 & $r\left(-1+\frac{P(t)}{\alpha}\right)\left(1-\frac{P(t)}{K_0}\right)P(t)+\frac{rP(t)}{t^{\alpha}}$ \\ \addlinespace

BPG11 & $rP(t)^{0.33}+rP(t)$ 
      & BPG23 & $r\left(1-\exp(-\gamma P(t))\right)P(t)+rP(t)\exp(-\gamma t)$ \\ \addlinespace

BPG12 & $r\left(1-\frac{P(t)}{K_0}\right)P(t)+rP(t)^{0.33}+rP(t)\exp(-\gamma t)$ 
      & BPG24 & $r\left(1-\frac{P(t)}{K_0}\right)P(t)+r\left(1-\exp(-\gamma P(t))\right)P(t)$ \\
\bottomrule
\end{tabular}
\end{table*}

\begin{table*}[htbp]
\centering
\small % 保持较小字号确保公式不拥挤
\newcolumntype{L}[1]{>{\raggedright\arraybackslash}p{#1}} % 定义支持自动换行且靠左对齐的列

\setlength{\tabcolsep}{4pt}       % 微调列间距以完美适配四列
\renewcommand{\arraystretch}{1.25} % 保持 1.25 的行高，给复杂的指数分式（如 Q/RT）提供足够的呼吸空间

\caption{\textbf{Material-domain LSR-Synth equations used in this paper.} Each entry reports the corresponding stress law $\sigma(\epsilon,T)=g(\epsilon,T)$.}
\label{tab:appendix_material_equations}

% 左右列宽比例：0.09 + 0.38 + 0.09 + 0.38 = 0.94\textwidth，给边缘留出安全空间
\begin{tabular}{@{} L{0.09\textwidth} L{0.38\textwidth} L{0.09\textwidth} L{0.38\textwidth} @{}}
\toprule
ID & Right-hand side & ID & Right-hand side \\
\midrule

MatSci1  & $E_0\epsilon\left(-\alpha_T(T-T_0)+1\right)-\beta(T-T_0)+\epsilon^M\eta(T-T_0)$ 
         & MatSci14 & $-\beta(T-T_0)+\epsilon\eta\exp\left(-(T-T_0)^2\right)$ \\ \addlinespace

MatSci2  & $H\epsilon^3+K\epsilon^N\exp\left(-\frac{Q}{RT}\right)+\epsilon\eta\sin(T-T_0)$ 
         & MatSci15 & $-\beta(T-T_0)+\epsilon^M\eta(T-T_0)$ \\ \addlinespace

MatSci3  & $H\epsilon^3+\eta(T-T_0)\exp(-\epsilon)$ 
         & MatSci16 & $E_0\epsilon\left(-\alpha_T(T-T_0)+1\right)+\epsilon\eta\exp\left(-(T-T_0)^2\right)$ \\ \addlinespace

MatSci4  & $H\epsilon^3+K\epsilon^N\exp\left(-\frac{Q}{RT}\right)+\epsilon^3\eta(T-T_0)$ 
         & MatSci17 & $E_0\epsilon^2+\epsilon\eta(T-T_0)^2$ \\ \addlinespace

MatSci5  & $E_0\epsilon^2+\eta(T-T_0)\log(\epsilon+1)$ 
         & MatSci18 & $E_0\epsilon\left(-\alpha_T(T-T_0)+1\right)-\beta(T-T_0)+\eta(T-T_0)\log(\epsilon+1)$ \\ \addlinespace

MatSci6  & $E_0\epsilon\left(-\alpha_T(T-T_0)+1\right)+K\epsilon^N\exp\left(-\frac{Q}{RT}\right)+\epsilon^M\eta(T-T_0)$ 
         & MatSci19 & $H\epsilon^3+\eta(T-T_0)\sin(\epsilon)$ \\ \addlinespace

MatSci7  & $E_0\epsilon\left(-\alpha_T(T-T_0)+1\right)+\epsilon\eta(T-T_0)^2$ 
         & MatSci20 & $E_0\epsilon^2-\beta(T-T_0)+\epsilon^3\eta(T-T_0)$ \\ \addlinespace

MatSci8  & $H\epsilon^3-\beta(T-T_0)+\eta(T-T_0)\log(\epsilon+1)$ 
         & MatSci21 & $E_0\epsilon^2+\epsilon\eta\sin(T-T_0)$ \\ \addlinespace

MatSci9  & $E_0\epsilon\left(-\alpha_T(T-T_0)+1\right)+\epsilon^M\eta(T-T_0)$ 
         & MatSci22 & $K\epsilon^N\exp\left(-\frac{Q}{RT}\right)-\beta(T-T_0)+\eta(T-T_0)\log(\epsilon+1)$ \\ \addlinespace

MatSci10 & $H\epsilon^3-\beta(T-T_0)+\epsilon^3\eta(T-T_0)$ 
         & MatSci23 & $E_0\epsilon\left(-\alpha_T(T-T_0)+1\right)+H\epsilon^3+\eta(T-T_0)\sin(\epsilon)$ \\ \addlinespace

MatSci11 & $H\epsilon^3+K\epsilon^N\exp\left(-\frac{Q}{RT}\right)+\epsilon\eta(T-T_0)^2$ 
         & MatSci24 & $K\epsilon^N\exp\left(-\frac{Q}{RT}\right)+\epsilon\eta\sin(T-T_0)$ \\ \addlinespace

MatSci12 & $K\epsilon^N\exp\left(-\frac{Q}{RT}\right)+\epsilon^3\eta(T-T_0)$ 
         & MatSci25 & $E_0\epsilon^2+E_0\epsilon\left(-\alpha_T(T-T_0)+1\right)+\eta(T-T_0)\log(\epsilon+1)$ \\ \addlinespace

MatSci13 & $E_0\epsilon\left(-\alpha_T(T-T_0)+1\right)+K\epsilon^N\exp\left(-\frac{Q}{RT}\right)+\epsilon\eta\exp\left(-(T-T_0)^2\right)$ 
         &          & \\ % 最后一组右侧留空，保持对齐
\bottomrule
\end{tabular}
\end{table*}

\section{Additional experimental results}

\subsection{Empirical budget--width trend}
\label{app:scaling_law}

We sweep PBeam and PIE over total budgets $N\in \{1,2,4,8,16,32,64,128\}$ and report means over three seeds after first averaging over all tasks within each domain. We fix $g=1$ when studying the width frontier. Because this analysis contains only eight discrete budget levels, we use it to support qualitative conclusions about compute allocation rather than to claim a universal scaling law.

\paragraph{The width frontier}
Whenever we analyze $w^\star(N)$, the frontier is defined as the best attainable score at a given pair $(N,w)$ after optimizing over all decompositions $(n,k)$ within PBeam and within PIE and then comparing the two algorithms. Therefore the optimal width is \emph{not} tied to a single control flow. The algorithm shown in Appendix Table~\ref{tab:appendix_width_optima} is simply the method that happens to attain that envelope at a given budget. When PBeam and PIE are theoretically identical, namely greedy search $(n,k)=(1,1)$ or single-step search $(T=1)$, missing PIE values are copied from PBeam. Based on this unified definition, the frontier lets us examine how the preferred width changes with budget after controlling for algorithm choice and the $(n,k)$ decomposition. The main conclusion is qualitative: the best width tends to increase with budget, although the exact argmax can be sensitive to finite-sample noise and nearby widths often perform similarly.

\begin{table*}[htbp]
    \centering
    \small
    \caption{\textbf{Exact optimal width on \texttt{gpt-oss-20b} without grouping.}
    For each pair $(N,w)$, we first optimize over all decompositions $(n,k)$ inside PBeam and inside PIE, and then choose the better algorithm. The table therefore reports which algorithm attains the envelope at each budget rather than a globally preferred control flow.}
    \begin{tabular}{ccccccc}
        \toprule
        & \multicolumn{3}{c}{Bio} & \multicolumn{3}{c}{Material} \\
        \cmidrule(lr){2-4}\cmidrule(lr){5-7}
        Budget $N$ & Algorithm & $(n,k,T)$ & $w^\star$ & Algorithm & $(n,k,T)$ & $w^\star$ \\
        \midrule
        1   & PBeam & $(1,1,1)$ & 1  & PBeam & $(1,1,1)$ & 1 \\
        2   & PBeam & $(1,1,2)$ & 1  & PBeam & $(1,1,2)$ & 1 \\
        4   & PBeam & $(1,2,2)$ & 2  & PBeam & $(2,1,2)$ & 2 \\
        8   & PIE   & $(1,2,4)$ & 2  & PIE   & $(2,2,2)$ & 4 \\
        16  & PIE   & $(1,2,8)$ & 2  & PBeam & $(1,4,4)$ & 4 \\
        32  & PBeam & $(1,8,4)$ & 8  & PBeam & $(1,4,8)$ & 4 \\
        64  & PBeam & $(2,8,4)$ & 16 & PIE   & $(8,2,4)$ & 16 \\
        128 & PBeam & $(4,8,4)$ & 32 & PIE   & $(64,1,2)$ & 64 \\
        \bottomrule
    \end{tabular}

    \label{tab:appendix_width_optima}
\end{table*}

\paragraph{Why we do not claim a universal width law.}
Although the optimal widths in \Cref{tab:appendix_width_optima} show a clear upward trend, the evidence is not sufficient to support a precise parametric law. First, the frontier contains only eight budget levels, all of which are powers of two and bounded by $N\le128$. Second, the optimal width is an argmax over noisy empirical measurements; when multiple widths are close, small seed-level fluctuations can change the selected width. Third, the trend may depend on verifier quality, model family, task difficulty, and the available parallel hardware.

For these reasons, we use the table as a practical allocation guide rather than as a scaling-law claim. A robust deployment strategy is to run a small pilot sweep over candidate widths, preferably powers of two, and then choose the smallest width whose performance is close to the observed frontier. This preserves most of the quality gains while avoiding over-interpretation of a small number of fitted points.

\paragraph{Performance under tuned width.}
When we select the best-performing width at each budget, training accuracy improves with additional compute but exhibits diminishing returns. We report this as a descriptive pattern rather than fitting a parametric power law, since the number of budget levels is small and the high-budget accuracies are already close to saturation. This result is still useful for compute allocation: most of the benefit comes from moving away from poorly matched width--depth regimes, while further gains at large budgets become progressively smaller.

\subsection{The study of second-order effect allocation choices}
\label{app:second_order}

\paragraph{Does the decomposition into $n$ and $k$ matter?}
Once the width $w=nk$ is fixed, the remaining question is how much of that width should come from expanding more parents ($n$) versus generating more children per parent ($k$). To isolate this effect, we fix the total budget to $N=128$, the global width to $w=32$, and grouping to $g=1$, and compare all admissible power-of-two decompositions of $32$. Within each domain and algorithm, we group these decompositions into three coarse regimes: \emph{population-heavy} ($n>k$), \emph{balanced} ($n\approx k$), and \emph{branching-heavy} ($k>n$). \Cref{tab:nk_assignment} reports the best $(n,k)$ within each regime.

The main pattern is that, after choosing the right width, the exact $(n,k)$ split is a second-order decision. Across the four domain--algorithm pairs, different regimes attain the best score, and most alternatives remain within about one percentage point of the regime optimum, with one larger outlier on Bio for PIE. This variation (typically $<1\%$) is much smaller than the cross-width effect reported in the main text: moving to a poorly chosen width can cost several percentage points even after optimizing over $(n,k)$ and the controller (can reach $\sim$5\%). In practice, this suggests a simple recipe: first tune the width $w$, then use a moderate decomposition such as balanced or population-heavy as a robust default rather than over-optimizing the exact $n$--$k$ split.

\begin{table*}[t]
  \centering
  \small
  \setlength{\tabcolsep}{6pt}
  \caption{\textbf{Best $(n,k)$ decomposition within each regime at fixed budget $N=128$, width $w=32$, and no grouping ($g=1$).}
  For each domain and algorithm, we partition the six admissible decompositions of $w=32$ into three regimes and report the best $(n,k)$ and its train $\operatorname{Acc}_{0.1}$ within each regime. Relative gap is computed against the best regime under the same domain and algorithm.}
  \label{tab:nk_assignment}
  \begin{tabular}{lllccc}
    \toprule
    Domain & Algorithm & Regime & Best $(n,k)$ & Train $\operatorname{Acc}_{0.1}$ & Rel.\ gap \\
    \midrule
    Bio & PBeam & Population-heavy & $(16,2)$ & 0.9816 & 0.28\% \\
    Bio & PBeam & \textbf{Balanced} & \textbf{$(4,8)$} & \textbf{0.9844} & \textbf{0.00\%} \\
    Bio & PBeam & Branching-heavy & $(1,32)$ & 0.9801 & 0.44\% \\
    \midrule
    Bio & PIE & \textbf{Population-heavy} & \textbf{$(16,2)$} & \textbf{0.9804} & \textbf{0.00\%} \\
    Bio & PIE & Balanced & $(8,4)$ & 0.9612 & 1.95\% \\
    Bio & PIE & Branching-heavy & $(2,16)$ & 0.9794 & 0.10\% \\
    \midrule
    Material & PBeam & Population-heavy & $(16,2)$ & 0.9857 & 1.15\% \\
    Material & PBeam & \textbf{Balanced} & \textbf{$(8,4)$} & \textbf{0.9971} & \textbf{0.00\%} \\
    Material & PBeam & Branching-heavy & $(1,32)$ & 0.9876 & 0.95\% \\
    \midrule
    Material & PIE & Population-heavy & $(32,1)$ & 0.9884 & 0.92\% \\
    Material & PIE & Balanced & $(4,8)$ & 0.9975 & $<0.01\%$ \\
    Material & PIE & \textbf{Branching-heavy} & \textbf{$(2,16)$} & \textbf{0.9976} & \textbf{0.00\%} \\
    \bottomrule
  \end{tabular}

  \vspace{2mm}
  \footnotesize
  Regime definitions:
  population-heavy $\{(32,1),(16,2)\}$,
  balanced $\{(8,4),(4,8)\}$,
  and branching-heavy $\{(2,16),(1,32)\}$.
\end{table*}

\paragraph{Does the choice between PBeam and PIE matter?}

\begin{figure*}[htbp]
    \centering
    \includegraphics[width=\textwidth]{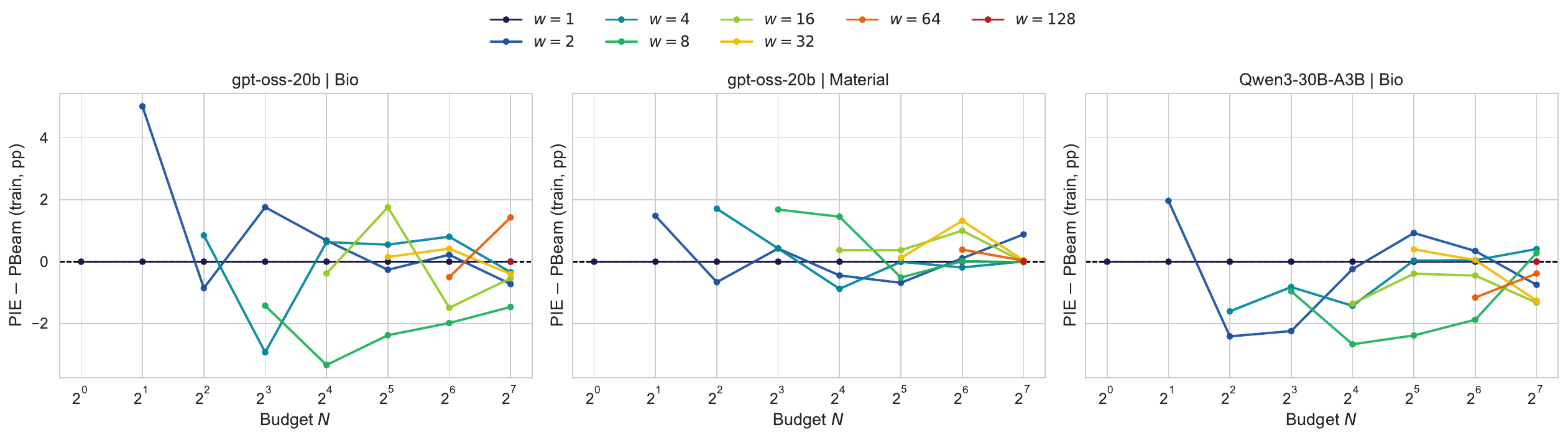}
    \caption{\textbf{PIE $-$ PBeam after optimizing $(n,k)$ inside each algorithm.}
    The vertical axis shows train $\operatorname{Acc}_{0.1}$ differences in percentage points. The effect size is much smaller than the effect of choosing the right width.}
    \label{fig:appendix_algo_delta}
\end{figure*}

We compare PIE and PBeam under the same budget $N$ and width $w$ with performance envelop over $(n,k)$ combination and plot the performance gap ($\text{PIE} - \text{PBeam}$) in Figure~\ref{fig:appendix_algo_delta}. The accuracy differences tightly fluctuate around the zero baseline, typically bounded within a marginal $\pm 2$ percentage points. This algorithmic variance can be safely ignored when compared to the 10 to 20 point penalty of choosing the wrong width. This confirms that PIE and PBeam are largely interchangeable, reinforcing our core thesis that the width $w$ determines the performance ceiling.

\begin{figure*}[htbp]
    \centering
    \includegraphics[width=0.96\textwidth]{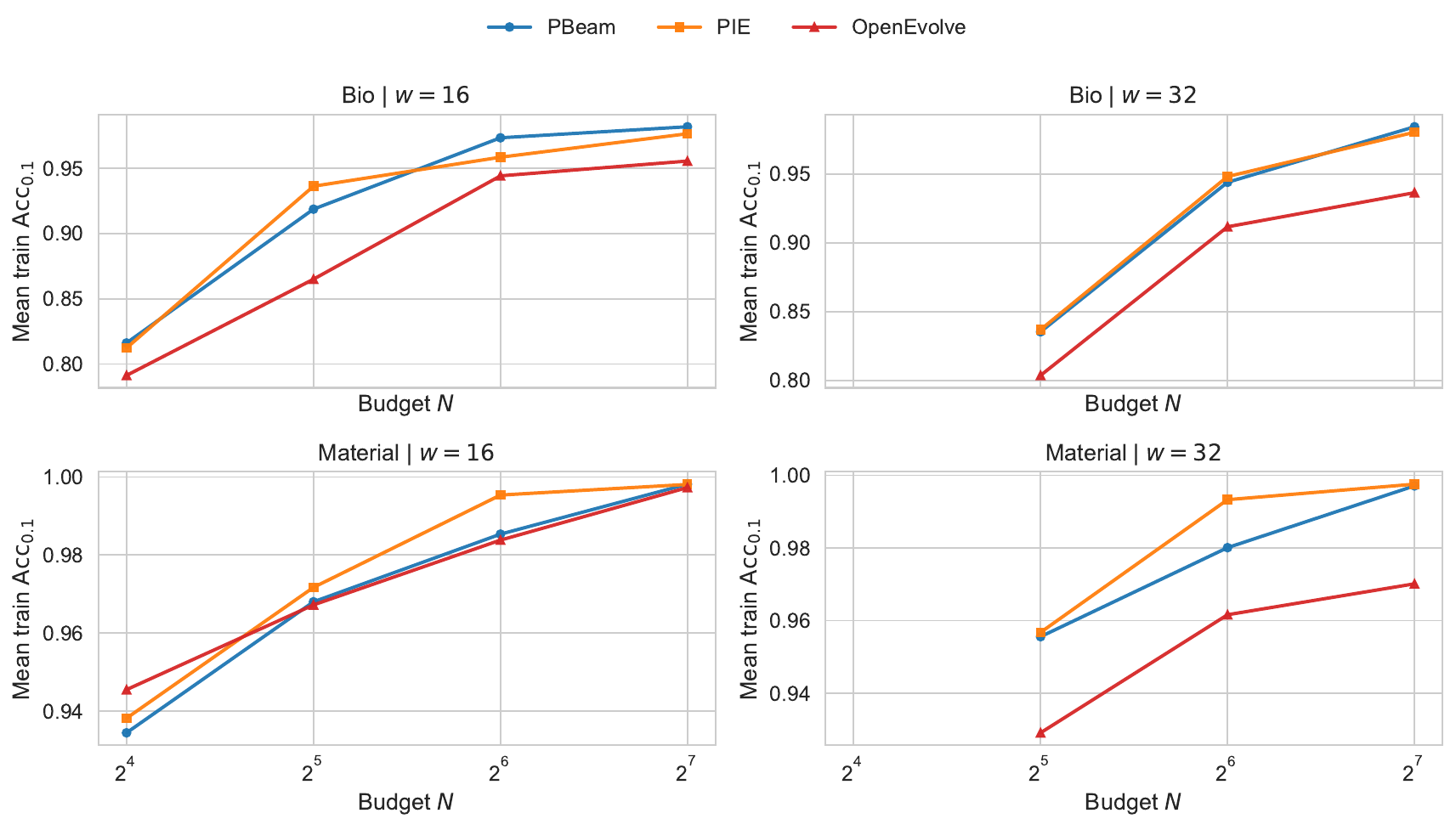}
    \caption{Matched comparison among PBeam, PIE, and OpenEvolve.}
    \label{fig:appendix_openevolve_compare}
\end{figure*}

\paragraph{Comparison with OpenEvolve.}
To isolate the effect of compute allocation from that of additional engineering heuristics, we compare PBeam and PIE against OpenEvolve under strictly matched global budgets. In the reduced $(n,k,g)$ parameterization introduced in the main text, OpenEvolve corresponds to a \emph{fully partitioned} special case:$
(n,k,g) = (w, 1, w)$,
so that each of the $g=w$ groups (islands) contains exactly one active parent, generates exactly one child per synchronized iteration, and therefore operates with local population $n/g=1$ and local width $w/g=1$. For a fixed global budget $N$ and width $w$, this gives OpenEvolve the same synchronized depth as our controllers,$T=\max\!\left(1,\left\lfloor \frac{N}{w}\right\rfloor\right),$
while differing only in how the width is organized within each iteration. This yields a clean matched-width comparison\footnote{We use the official implementation of OpenEvolve with the same prompt template in~\cref{appendix:benchmark}. Unless specially mentioned, the other hyperparameters are set as default values.} against PBeam and PIE, which can realize the same global width with less aggressive partitioning (i.e., smaller $g$ and/or $k>1$). As shown in \cref{fig:appendix_openevolve_compare}, PBeam and PIE generally achieve better train $\operatorname{Acc}_{0.1}$ than OpenEvolve at the same $(N,w)$ when using a mild $g=2$ and balanced $(n,k)$ assignment. This is consistent with our grouping analysis in \cref{sec:expr}: fully partitioning the search into $w$ isolated groups can reduce useful competition across candidates and push each group into an overly narrow local search regime. Overall, this comparison supports our main claim that, on this benchmark, the dominant gains come from getting the global compute allocation right, whereas additional heuristics such as island engineering, crossover, or specialized prompting appear to be secondary.

\subsection{Wall-clock time speedups through grouping}
\label{app:wall-clock}

As claimed in the main paper, grouping improves inference efficiency via reduced synchronization overhead. Intuitively, one might hypothesize that evaluating candidates in larger groups could artificially enforce search diversity---by allowing parallel branches to explore independently before being evaluated and pruned---which might in turn elevate the overall reasoning performance. However, our empirical results reveal that this is generally not the case: grouping predominantly acts as a strict accuracy-efficiency trade-off rather than an algorithmic performance booster.

\begin{figure*}[t]
    \centering
    \includegraphics[width=\textwidth]{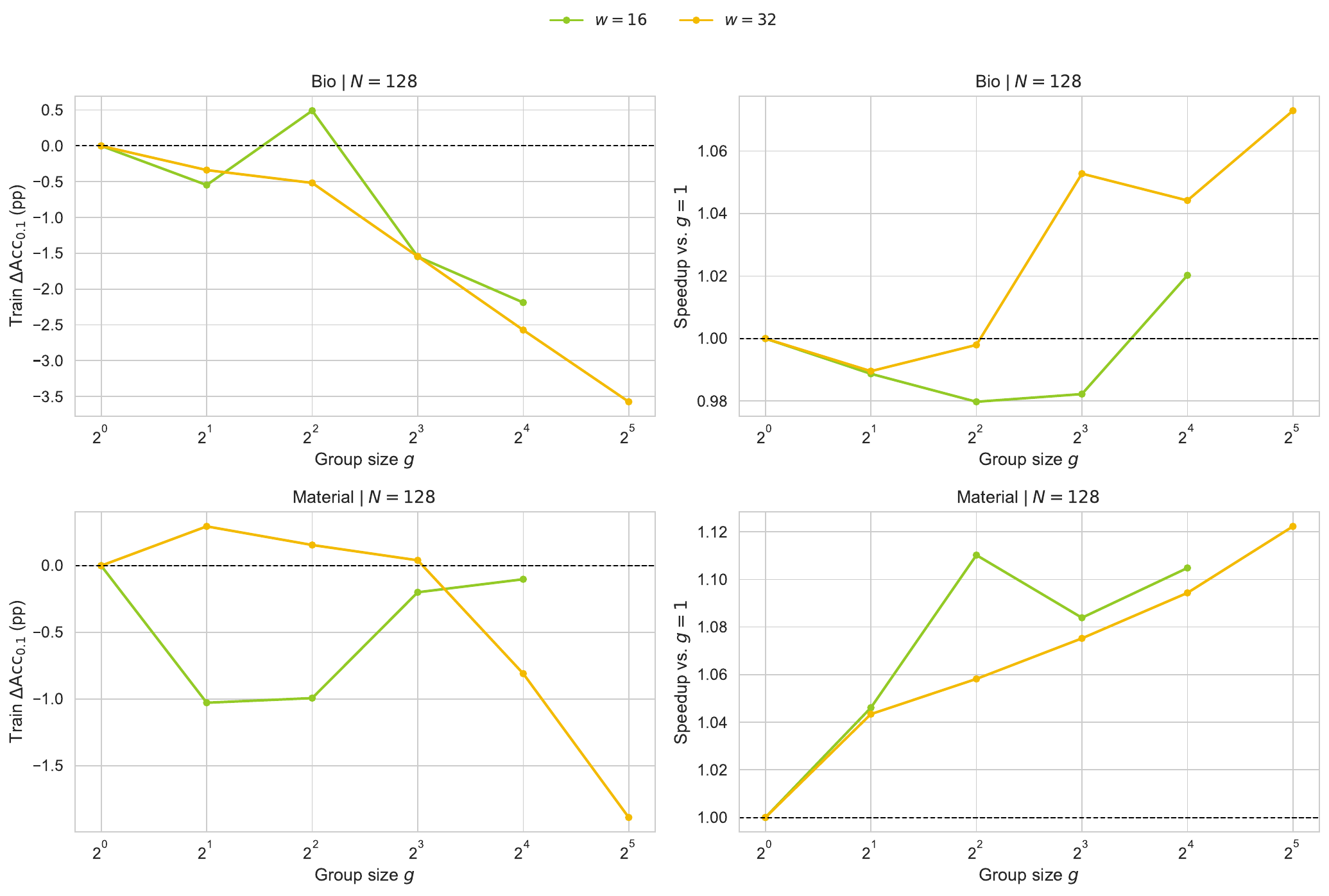}
    \caption{\textbf{Effect of group size on performance and wall-clock time.}
    For each domain and width, every point is the best measured configuration at fixed $(w,g)$ and $N=128$, compared against the best ungrouped configuration with the same width.
    Moderate grouping can improve throughput with modest quality loss, but excessively large $g$ eventually collapses the per-group width and hurts performance.}
    \label{fig:appendix_group_tradeoff}
\end{figure*}

To isolate the exact impact of grouping once the search width is already chosen, Figure~\ref{fig:appendix_group_tradeoff} fixes the compute budget at $N=128$ and compares the optimal grouped configurations ($g > 1$) against their exact ungrouped baselines ($g=1$) of the same width.

{On the Material domain, grouping can occasionally be helpful: at width $16$, moving from $g=1$ to $g=2$ increases the train $\operatorname{Acc}_{0.1}$ by $0.295$ percentage points and yields a $1.043\times$ speedup, while $g=8$ still preserves a slight accuracy gain alongside a $1.075\times$ speedup. In contrast, the Bio domain is more fragile. At width $16$, setting $g=8$ provides a $1.053\times$ speedup but suffers a severe accuracy penalty, losing $1.54$ percentage points. Ultimately, while grouping delivers measurable system-level acceleration, overly large $g$ should be used cautiously.}

\subsection{Generalization: internal train--test gap and cross-model consistency}
\label{app:train-test}

\begin{figure*}[htbp]
    \centering
    \includegraphics[width=\textwidth]{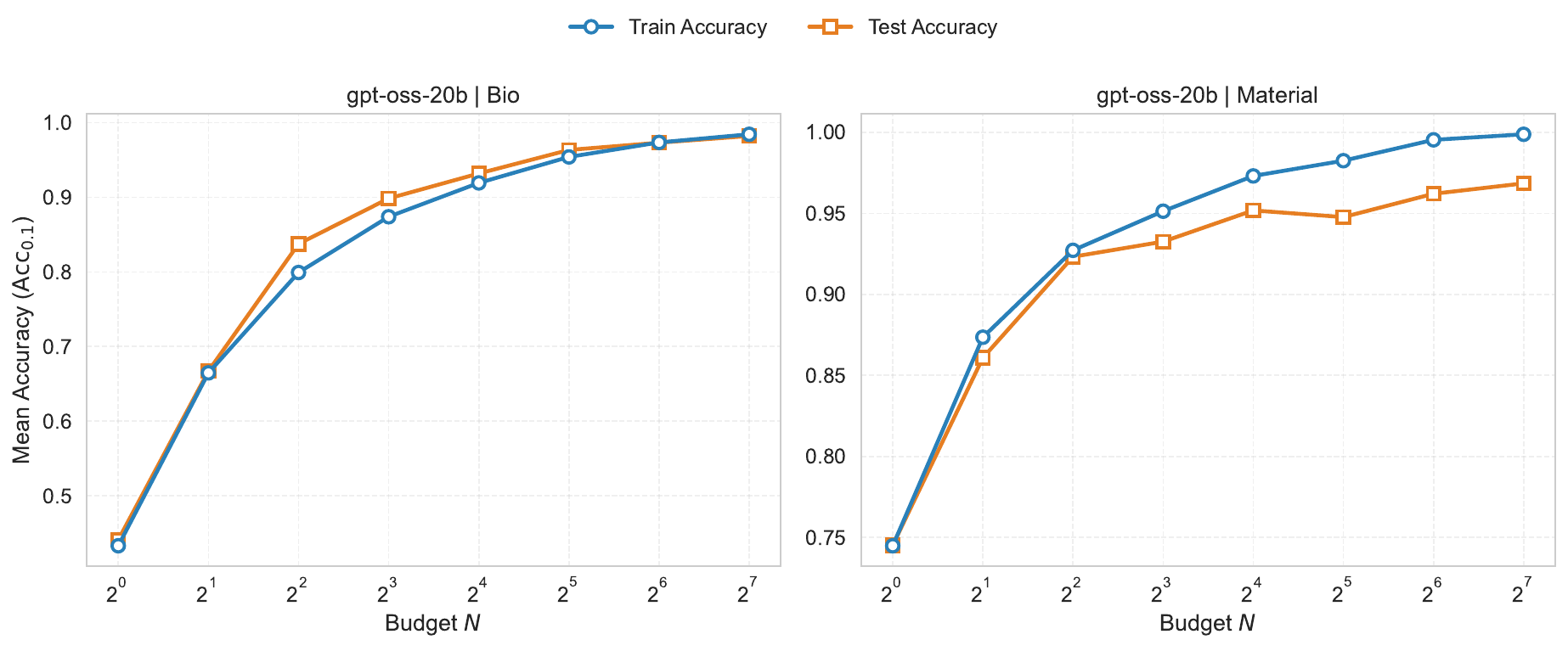}
    \caption{\textbf{Train vs.\ test performance of gpt-oss-20b.}
    Mean accuracy ($\operatorname{Acc}_{0.1}$) on the training and test sets for the Bio (left) and Material (right) domains across varying compute budgets $N$.}
    \label{fig:appendix_train_test_gap}
\end{figure*}

\paragraph{Generalization from training to test sets}
As discussed in the main paper, we determine the search configuration (e.g., the search width $w^\star$) using training-set performance. As shown in Figure~\ref{fig:appendix_train_test_gap}, the test accuracy closely tracks the training accuracy across all compute budgets, indicating no obvious overfitting. However, optimizing strictly on the training set at larger budgets can occasionally lead to a slight performance drop on the unseen test set (e.g., $N \ge 2^5$ in the Material domain). Furthermore, our control flow is designed to greedily maximize the training accuracy at each iteration and strictly retain the highest-scoring paths during each step of the beam search. While this aggressive, step-wise optimization guarantees peak performance on the training distribution, it inherently risks mild overfitting. Consequently, the resulting configurations may yield slightly sub-optimal accuracy on the unseen test set.

\begin{figure*}[h]
    \centering
    \includegraphics[width=0.6\textwidth]{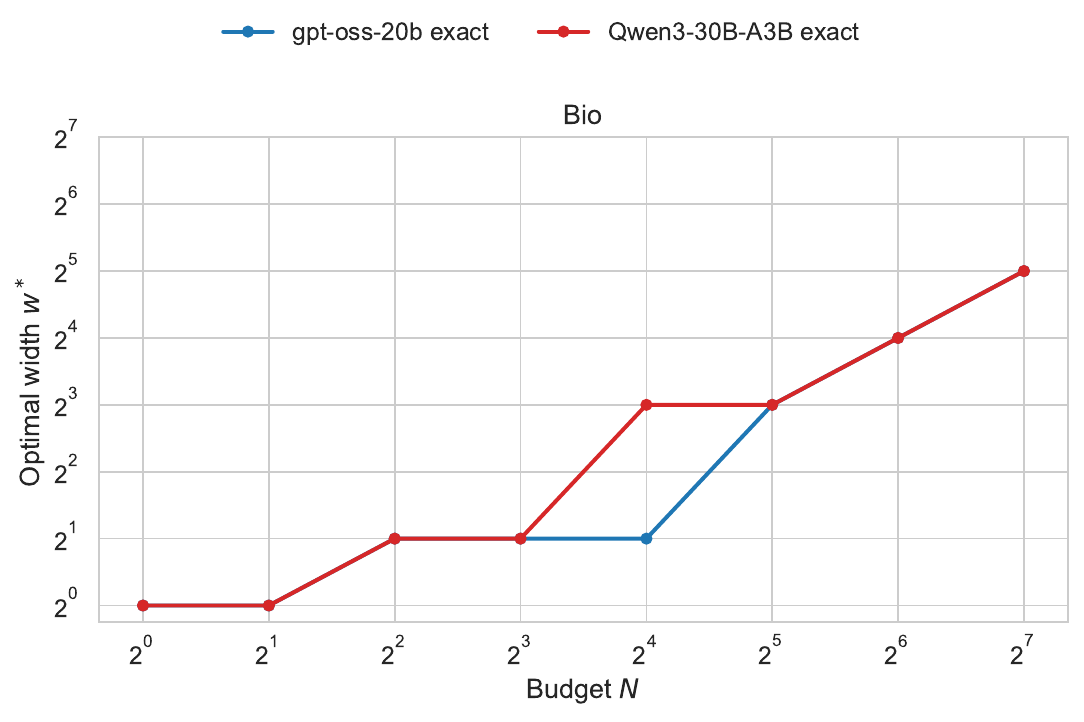}
    \caption{\textbf{Auxiliary cross-model comparison on Bio with \texttt{Qwen3-30B-A3B}.}
Solid lines show the exact best-performing width in the sweep. The same qualitative width--depth trade-off appears across models, although the exact mid-budget operating point remains backbone-dependent.}
    \label{fig:appendix_cross_model}
\end{figure*}

\paragraph{Does the optimal width law generalize across backbones?} 
To investigate whether our empirically derived width scaling behavior is tied to a specific model architecture, we evaluate the \texttt{Qwen3-30B-A3B} backbone on the Bio domain. As illustrated in Figure~\ref{fig:appendix_cross_model}, the qualitative width--budget trend is similar across the two model families.
 The exact optimal width trajectories for both \texttt{Qwen3} and \texttt{gpt-oss} perfectly overlap at early compute budgets ($N \le 8$). The only localized divergence occurs at the intermediate budget of $N=16$, where Qwen temporarily favors a deeper search ($w^\star=2$) while the GPT backbone expands to a wider allocation ($w^\star=8$). This aligned trajectory suggests that the width--budget trade-off is not unique to one backbone, although the exact preferred width remains model- and budget-dependent.
% Crucially, the two curves perfectly reconverge for all budgets from $N=32$ onward. Furthermore, their relaxed frontiers (the smallest widths yielding performance within $0.1$ percentage points of the absolute best) track the exact same broad log-linear trend.

\clearpage

\end{document}